%% file: main.tex
\documentclass{article} 

\usepackage{fullpage}
\RequirePackage{natbib}

\setcitestyle{authoryear,round,citesep={;},aysep={,},yysep={;}}

\input{include/math_commands}

\usepackage[utf8]{inputenc} 
\usepackage[T1]{fontenc}    
\usepackage[font=normalsize,labelfont=bf]{caption}

\usepackage[backref=page,colorlinks=true,linkcolor=blue,citecolor=green,urlcolor=blue]{hyperref}
\usepackage{url}            
\usepackage{booktabs}       
\usepackage{amsfonts}       
\usepackage{nicefrac}       
\usepackage{microtype}      
\usepackage{xcolor}         
\usepackage{graphicx}
\usepackage{algorithm}
\usepackage[noend]{algorithmic} 
\usepackage{wrapfig}
\usepackage{amsmath}
\usepackage{bbm}
\usepackage{bm}
\usepackage{subcaption}
\usepackage{soul}
\usepackage[normalem]{ulem}
\usepackage{floatrow}
\DeclareFloatSeparators{hfill}{\hfill}
\usepackage{xcolor}

\newtheorem{theorem}{Theorem}[section]

\usepackage{array}
\usepackage{sidecap}
\usepackage{comment}
\usepackage{colortbl}
\usepackage{multirow}
\usepackage{ulem}
\usepackage{enumitem}
\setlist[enumerate]{leftmargin=5mm, label=\alph*)}
\setlist[itemize]{leftmargin=5mm}

\title{\LARGE\textbf{EnsemW2S: Enhancing Weak-to-Strong Generalization with Large Language Model Ensembles}} 
\date{}

\author{
  Aakriti Agrawal\thanks{University of Maryland; e-mail: {\tt \{agrawal5, furongh\}@umd.edu}}
  \and
  Mucong Ding\footnotemark[1]
  \and
  Zora Che\footnotemark[1]
  \and
  Chenghao Deng\footnotemark[1]
  \and
  Anirudh Satheesh\footnotemark[1]
  \and
  Bang An\footnotemark[1]
  \and
  Bayan Bruss\footnotemark\thanks{Capital One}
  \and
  John Langford\thanks{Microsoft}
  \and
  Furong Huang\footnotemark[1] \footnotemark[3]
}

\begin{document}

\maketitle

\input{sections/0-abstract}
\input{sections/1-Intro}
\input{sections/2-Adaboost}
\input{sections/3-Method}
\input{sections/RelatedWorks}
\input{sections/4-Experiment_Setup}
\input{sections/5-Results}

\input{sections/6-Conclusion}

\newpage
\section{Acknowledgement}

Agrawal, Ding, Deng, Che, Satheesh, An and Huang are supported by DARPA Transfer from Imprecise and Abstract Models to Autonomous Technologies (TIAMAT) 80321, National Science Foundation NSF-IIS-2147276 FAI, DOD-AFOSR-Air Force Office of Scientific Research under award number FA9550-23-1-0048, Adobe, Capital One and JP Morgan faculty fellowships.
\bibliography{references}
\bibliographystyle{include/icml2025}


\newpage
\appendix
\input{sections/Appendix}

\end{document}

%% file: include/math_commands.tex
\usepackage{amsmath,amsfonts,bm}
\usepackage{floatrow}

\def\eqref#1{equation~\ref{#1}}

\def\1{\bm{1}}

\DeclareMathAlphabet{\mathsfit}{\encodingdefault}{\sfdefault}{m}{sl}
\SetMathAlphabet{\mathsfit}{bold}{\encodingdefault}{\sfdefault}{bx}{n}



%% file: sections/0-abstract.tex
\begin{abstract}
With Large Language Models (LLMs) rapidly approaching and potentially surpassing human-level performance, it has become imperative to develop approaches capable of effectively supervising and enhancing these powerful models using smaller, human-level models exposed to only human-level data. We address this critical weak-to-strong (W2S) generalization challenge by proposing a novel method aimed at improving weak experts, by training on the same limited human-level data, enabling them to generalize to complex, super-human-level tasks. Our approach, called \textbf{EnsemW2S}, employs a token-level ensemble strategy that iteratively combines multiple weak experts, systematically addressing the shortcomings identified in preceding iterations. By continuously refining these weak models, we significantly enhance their collective ability to supervise stronger student models. We extensively evaluate the generalization performance of both the ensemble of weak experts and the subsequent strong student model across in-distribution (ID) and out-of-distribution (OOD) datasets. For OOD, we specifically introduce question difficulty as an additional dimension for defining distributional shifts. Our empirical results demonstrate notable improvements, achieving 4\%, and 3.2\% improvements on ID datasets and, upto 6\% and 2.28\% on OOD datasets for experts and student models respectively, underscoring the effectiveness of our proposed method in advancing W2S generalization.

\end{abstract}

%% file: sections/1-Intro.tex
\section{Introduction}\label{sec:intro}
Rapid advancements in Large Language Models (LLMs) have largely been driven by extensive internet-scale data and improvements in computational hardware.  However, these vast datasets have largely reached their saturation point, offering diminishing returns in terms of new knowledge, and hardware-driven performance gains are also diminishing. Despite these limitations, the need for LLMs to effectively generalize to out-of-distribution, challenging, and unseen data remains crucial.  This capability becomes even more essential as models approach and potentially exceed superhuman performance \cite{openAIw2s}, necessitating super-human level supervision from human-level data and models. In this paper, we focus on enhancing the generalization performance of human-level experts using limited human-level labeled data and smaller-scale models. We subsequently leverage these improved models to supervise and enhance larger-scale models on superhuman-level data. Our framework is inspired from the ``superalignment'' or ``w2s generalization'' problem by \citet{openAIw2s}, but extends it by also focusing on generalization to superhuman-level data using only human-level labeled data.


W2S generalization has been previously studied \cite{sun2024easy, openAIw2s, aligner, quant_w2s, theory_w2s}; however, several key challenges remain unresolved: \textbf{(C1) Generalization to Out-of-distribution (OOD) and super-human level data.} 
One key aspect of W2S generalization is the ability to extrapolate from human-level performance to super-human tasks using only easily accessible human-level data. While ``easy'' examples are abundant and cheap to label, ``hard'' or super-human examples often lie beyond the reach of annotators, either due to cognitive limits or prohibitive costs. In fact, for many tasks, groundtruth labels at the superhuman level may be \textit{fundamentally unavailable}---raising a critical obstacle to benchmarking generalization at that level. Therefore, we believe that easy-to-hard (E2H) generalization is central to the W2S paradigm. To address this, we explicitly partition data into ``easy'' and ``hard'' categories and develop methods that promote generalization to harder questions even when trained solely on easy examples.
\textbf{(C2) Limitation of Single Weak Supervisors.} Previous research predominantly utilizes a single weak supervisor, often leading to biased, inconsistent, or incomplete supervisory signals. In practice, multiple complementary weak experts (e.g., humans specializing in different domains) can be readily accessed. Our work leverages an ensemble approach that aggregates these diverse weak supervisors, providing a richer, more balanced supervisory signal and significantly enhancing overall model generalization. \textbf{(C3) Lack of Iterative Weak Model Enhancement.} Existing approaches typically treat weak models as static entities, neglecting the potential for iterative improvements through limited labeled data. Although methods such as voting \citep{anonymous2025collab} and debate \citep{du2023improving} have attempted model combination during inference, they require a stronger supervisor (reward model or judge) to select optimal tokens or answers. In contrast, our approach iteratively refines weak models without relying on stronger supervisory signals or additional data, thereby fostering a self-reinforcing cycle of continuous improvement. This dynamic enhancement not only promotes continuous improvement of weak models for improved w2s generalization but also holds potential for answering previously unsolved and challenging questions. 

\begin{figure}[t]
    \centering
    \includegraphics[width=\linewidth, trim={0 20 0 0}, clip]{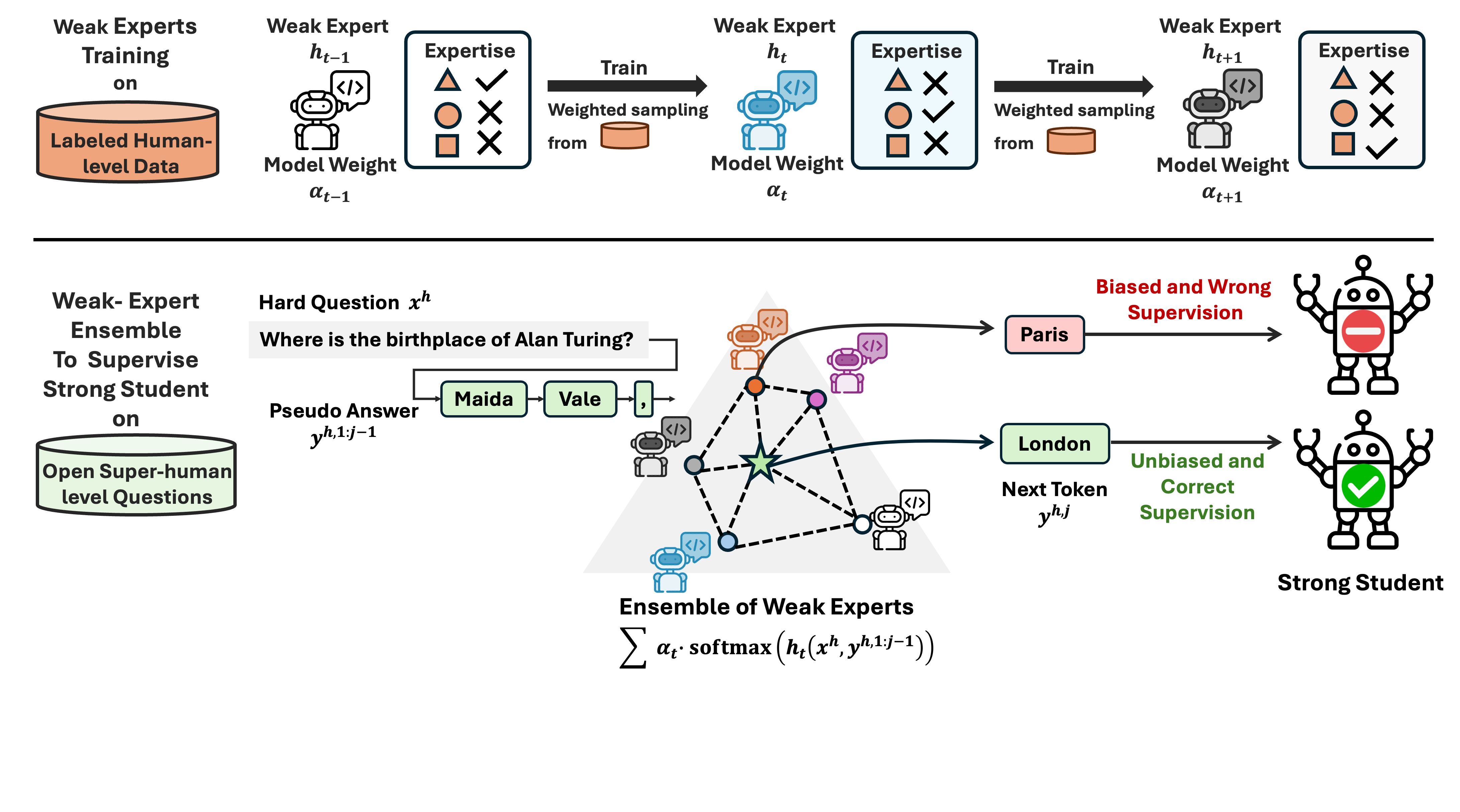}
    \vspace{-4em}
    \caption{This figure illustrates the full pipeline of EnsemW2S. The top part shows how weak experts are iteratively improved by training on the same human-level labeled data. The bottom part depicts how these experts are combined at the token level during inference to supervise the strong student model. Training data is sampled using token-level weights, which define a distribution over training tokens. Based on the token-level errors of each trained weak model, we compute model weights ($\alpha_t$), which are then used during inference to compute a weighted combination of token-level probabilities. This aggregated distribution is used to generate answer tokens for previously unseen, superhuman-level questions, providing supervision to the strong model—following the setup in \citet{openAIw2s}. EnsemW2S reflects a realistic scenario where weak experts perform well on easy questions and are leveraged to supervise a stronger model on harder (and potentially unsolved) tasks.} 
    \label{fig:method}
\end{figure}

To address the above challenges, we propose \textbf{EnsemW2S}, a novel token-level ensemble algorithm for autoregressive LLMs. Inspired by the spirit of AdaBoost---an ensemble method for binary classifiers known for improving generalization while resisting overfitting---EnsemW2S trains and combines complementary weak models at the token level.
Specifically, it adjusts token probabilities in a manner analogous to controlled decoding methods \citep{mudgal2023controlled}, but \textit{extends them to a multi-model setting}. The approach iteratively improves newly introduced weak models based on identified shortcomings of prior models without requiring any additional labeled data. 

We evaluate generalization rigorously on unseen data, distinguishing between in-distribution (ID) and OOD scenarios defined either by different datasets but the same task or question difficulty levels. 
We refer to this as \textbf{easy-to-hard (E2H) generalization}, a proxy for the unattainable goal of generalizing from human-level to superhuman performance---where groundtruth labels are inherently unavailable.
By leveraging a small held-out labeled set for calibration, EnsemW2S enables effective supervision that improves both weak and strong models. 
As a result, the final strong model attains significantly better generalization and can even match or outperform models trained on full groundtruth datasets.

The \textbf{main contributions} of this paper are the following:\\
\textbf{(1) EnsemW2S: A novel \textit{token-level ensemble method}} to improve generalization to difficult data and further improve w2s generalization. \textit{(a)} EnsemW2S \textit{iteratively improves weak models} without using additional data and aggregates them via a probabilistic voting mechanism. This self-reinforcing cycle of improvement significantly enhances generalization. \textit{(b)} We also provide \textit{theoretical justification} for exponential decay in training error bounds as more weak models are incorporated (assuming our new weak learning condition holds). We then discuss how training data margin distribution improves, leading to better generalization while preventing overfitting. \textit{(c)} We achieve \textit{improved generalization to unseen data by \textbf{upto 4\% for ID and 4.34\%, 6\% for OOD}} (easy-hard) with test data from the same dataset and a different dataset, respectively.\\
\textbf{(2) Better weak models further improve W2S Generalization} by providing more effective supervision. We achieve \textit{performance improvement of up to 3.2\% on ID and 2.42\%, 2.28\% on OOD  (easy-hard) with test data from the same dataset and a different dataset}, respectively.

%% file: sections/2-Adaboost.tex
\section{Framework}\label{sec:w2sviae2h}
\begin{figure}[htp]
  \centering
  \begin{minipage}{0.55\textwidth}
    \includegraphics[width=\linewidth, trim={15 5 20 5}, clip]{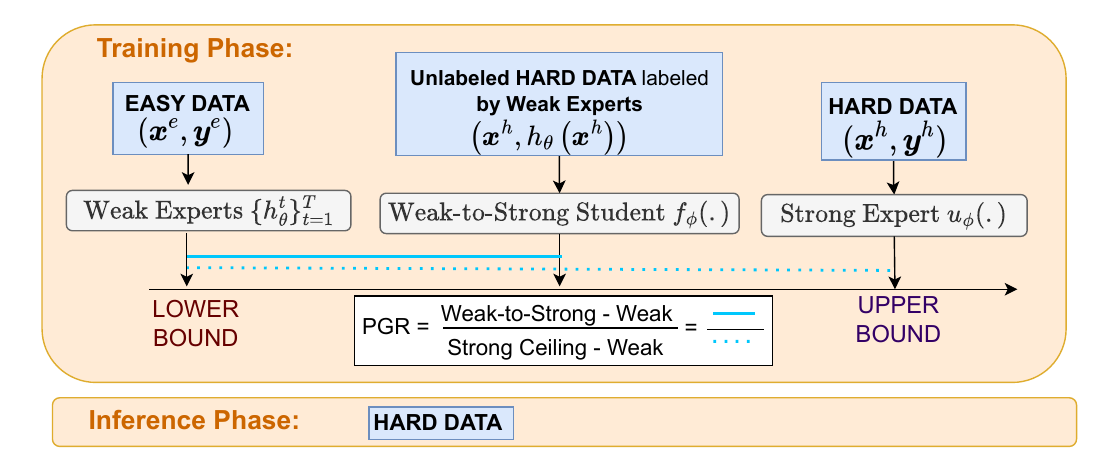}
  \end{minipage}%
  \hfill
  \begin{minipage}{0.42\textwidth}
  \captionof{figure}{Overview of the EnsemW2S. EnsemW2S leverages hard examples with pseudo-labels generated by weak experts to train a strong student model. To evaluate its effectiveness, we measure Performance Gap Recovered (PGR)~\citep{openAIw2s}}
    \label{fig:train_flow}
  \end{minipage}
\end{figure}

\paragraph{W2S Generalization with E2H Generalization}
As a proxy for human and superhuman-level data, we split data into easy and hard, using the former to generalize to the latter. This E2H framework is a more pragmatic setting to study the (im)possibility of w2s generalization. in this framework, weak models (or human-level models) trained on simpler tasks guide a strong model (or superhuman-level model) toward solving more complex challenges, mirroring real-world conditions of limited human supervision. Figure~\ref{fig:train_flow} outlines EnsemW2S, where \textbf{weak expert models ($h_\theta$)} proficient in `easy tasks' $(\bm{x}^e,\bm{y}^e)$ supervise a \textbf{strong model ($f_{\phi}$)} to solve (unsolvable) hard problems ($\bm{x}^h$). The strong student model under EnsemW2S is trained using unlabeled hard data alongside pseudo-labels, $(\bm{x}^h, h_{\theta}(\bm{x}^h))$, where $h_{\theta}(\bm{x}^h)=\text{func}(\{h_{\theta}^t\}_{t=1}^T,\bm{x}^h)$ denotes the ensemble-generated labels from weak experts. Ultimately, our goal is to improve the \textbf{Performance Gap Recovered (PGR)} which compares EnsemW2S against two baselines: a lower bound---weak experts trained on easy data and evaluated on hard data---and an upper bound---the performance of a strong model trained directly on hard data. The upper bound or oracle, is the \textbf{strong model ($u_{\phi}$)}, which matches the student model in size and caliber but is trained using ground-truth labeled hard data $(\bm{x}^h, \bm{y}^h)$, typically unattainable in real-world scenarios. This oracle sets the upper performance limit for W2S models.


%% file: sections/3-Method.tex
\section{EnsemW2S: Improving W2S via Improved Expert Ensembles}\label{sec:w2sviaadaboost} 
In this section, we introduce \textbf{EnsemW2S}, a token-level ensemble method for autoregressive generative models such as LLMs. Figure~\ref{fig:method} outlines the method. As classification is a subset of generation, we first discuss ensemble methods for binary classification LLMs, followed by the challenges of applying such methods to generation tasks, and finally present our proposed solution.
\subsection{Motivation: AdaBoost for Binary Classification with LLMs}
AdaBoost is a well-known ensemble learning algorithm designed for binary classification. We apply AdaBoost to a binary classification LLM to study W2S generalization. Historically, AdaBoost demonstrates strong generalization on unseen data by increasing the margins of training examples, which correlates with lower generalization error, even after achieving zero training error-making it ideal for this study. AdaBoost iteratively reweights training examples, and trains weak experts one at a time focusing on hard-to-classify samples (check Line 5 of Algorithm \ref{alg:adaboost}). The essential requirement is that each weak learner must perform slightly better than random guessing, thereby satisfying the weak learning condition. The outputs of these weak experts are aggregated through a weighted majority vote to produce pseudo-labels: $\mathbbm{1}\left(\sum_{t=1}^T \alpha_t h_{\theta}^t (\bm{x}^h)>0\right)\in {0,1}$, where $\alpha_t$ are hyperparameters weighing each weak expert based on accuracy. For a detailed mathematical summary, refer to Appendix~\ref{appx:adaboost}. We then use this ensemble to generate answers for challenging (``hard'') questions, $\bm{x}^h$.

\subsection{Challenges of Ensemble for Generation Task} 
Popular ensemble methods like AdaBoost \cite{boost} leverage the “wisdom of the crowd” to build stronger classifiers by combining weaker ones, typically through classification scores. Inspired by AdaBoost’s ability to increase margin and improve generalization, we explore whether its principles can be extended to generation tasks. However, applying AdaBoost to autoregressive LLMs is infeasible due to several key challenges: (1) \textbf{Single-Class vs Multi-Class:} Unlike binary classification, each token prediction in LLMs is a multi-class problem over a large vocabulary (typically >4000 classes). Aggregating such high-dimensional probability distributions introduces significant computational overhead and potential numerical instability. (2) \textbf{Autoregressive Generation:} Unlike classification, where scores are combined over a fixed output space, generation tasks involve sequential prediction, where early token errors can propagate and affect future outputs. This makes ensemble aggregation over generations inherently more complex. (3) \textbf{Variable-Length Outputs:} LLMs generate variable-length outputs, making it non-trivial to combine their responses consistently.


\subsection{EnsemW2S: LLM Token-Level Ensemble Method}
To address the above challenges, we propose a multi-class, generation based, AdaBoost inspired, LLM ensemble algorithm where the number of classes corresponds to the vocabulary size. To develop this algorithm, we treat each token as an independent sample and include error for each token. Algorithm \ref{alg:multiadaboost} provides the steps for EnsemW2S algorithm for Multiple-Choice Q/A tasks. In the sections below we provide detail of each step along with theoretical proofs. Appendix Fig. \ref{fig:train_flow_detail} gives details on algorithmic and data flow. Note, AdaBoost is a special case of EnsemW2S for binary classification task (see Algorithm \ref{alg:adaboost} in Appendix \ref{appx:adaboost}). Refer Figure \ref{fig:method} for outline and intuition behind our method.

\textbf{Token-Level Weighting.} Our method 
generates weights for each token within a sentence sample. We define the initial token-sample weights vector $D_1(i, j) \leftarrow \frac{1}{n}$ for all $i \in [m], j \in [k_{i}]$, where $n = \sum_{i=1}^{m} k_{i}$, $k_{i}$ is the number of tokens in the answer part of each sample $i$, $m$ is the total number of training data samples and $j$ is the $j^\text{th}$ token in a particular chosen $i^\text{th}$ sample. We update these weights, $D_{t}(i,j)$, for each iteration $t$ of EnsemW2S. Note here that \textbf{unlike AdaBoost}, EnsemW2S assigns weights to all tokens in a sample, adjusting for varying lengths, whereas AdaBoost maintains only one weight for each sample.

\textbf{Token-Level Data Sampling.} 
We sample $S' = \{(\bm{x'}^{e}_i, \bm{y'}^{e}_i)\}_{i=1}^m$ from $S$ using token-sample weights $D_{t}(i,j)$. By sampling with respect to probability masses $D_{t}(i,j)$ with repetition, we obtain a set of $n = \sum_{i=1}^{m} k_{i}$ tokens to train on. However, treating these $n$ sampled tokens as independent training samples is very inefficient. Instead, we ``assemble'' the sampled tokens back into the sentences they belong to and implement label masking to only train on the sampled tokens in each sentence. Following \textbf{this novel treatment proposed}, we can train on sampled tokens with minimal overheads.

\textbf{Training and Generating New Weak Experts.} For each iteration, $t$, of EnsemW2S algorithm we train a new weak expert model $h^t_{\theta}$ on the sampled data, $S'$. In this way, EnsemW2S \textit{iteratively improves weak models} and recruits additional weak models throughout the training process---\textbf{the first approach} to dynamically refine weak models while training the strong model. This self-reinforcing cycle of improvement significantly enhances generalization.

\begin{algorithm}[!ht]
\caption{Main Algorithm: EnsemW2S}
\label{alg:multiadaboost}
\begin{algorithmic}[1]
\REQUIRE ``Easy'' Q/A dataset \( S^e = \{(x_i^e, y_i^e)\}_{i=1}^{m} \), pre-trained weak expert model \( h_\theta^0 \), iterations \( T \), ``Hard'' unlabeled (questions only) dataset \( S^h = \{x_o^h\}_{o=1}^{O} \)
\ENSURE Weak-to-Strong Student Model \( f_{\phi}(\cdot) \)
\STATE \textbf{Initialize Token-Sample Weights:} \\
$ D_1(i, j) \gets \frac{1}{n}, \quad \forall i \in [m], j \in [k_i] $, where $k_{i}$ is token length of $\bm{y}^{e}_i = (\bm{y}_i^{e, 1}, \bm{y}_i^{e, 2} ... \bm{y}_i^{e, k_{i}})$ and $n = \sum_{i=1}^{m} k_{i}$ 
\STATE \textbf{Compute Pre-Training Error} of $h_{\theta}^0$:
$\epsilon_{\text{pre}} \gets \sum_{i=1}^{m} \sum_{j=1}^{k_i} \mathbbm{1} \{ h_\theta^0(x_i^e, y_i^{e,j-1}) \neq y_i^{e,j} \} D_1(i, j)$
\FOR{$t \gets 1$ to $T$}
    \STATE Sample $ S' = \{(x_i^e, y_i^{e'})\}_{i=1}^{m} $ from $ S^e $ using $ D_t(i, j) $
    \STATE Train new weak expert model $ h_\theta^t $ on $ S' $
    \STATE Compute error:
    $\epsilon_t \gets \sum_{i=1}^{m} \sum_{j=1}^{k_i} \mathbbm{1} \{ h_\theta^t(x_i^e, y_i^{e,j-1}) \neq y_i^{e,j} \} D_t(i, j)$
    \IF{$\epsilon_t \geq \epsilon_{\text{pre}}$}
        \STATE \textbf{Break}
    \ENDIF
    \STATE Compute model weight:
    $ \alpha_t \gets \log \frac{1 - \epsilon_t}{\epsilon_t} + \log \left(\frac{1}{1 - \epsilon_{\text{pre}}} - 1\right) $
    \STATE Update sample weights: 
    $ D_{t+1}(i, j) \gets \frac{D_t(i, j)}{Z_t} e^{\alpha_t \mathbbm{1} \{ h_\theta^t(x_i^e, y_i^{e,j-1}) \neq y_i^{e,j} \} } \forall i \in [m], j \in [k_{i}] $, where  $Z_t = \sum_{i=1}^{m} \sum_{j=i}^{k_i} D_t(i,j) \cdot e^{\alpha_t \mathbbm{1} (h_{\theta}^t ( \bm{x}^{e}_i, \bm{y}_{i}^{e, j-1}) \neq \bm{y}_{i}^{e, j} )}$
\ENDFOR
\FOR{$o \gets 1$ to $O$}
    \FOR{$j \gets 1$ to $k_o$}
        \STATE Autoregressively generate the $j^\text{th}$ token of the ``pseudo-answer''
        \setlength{\abovedisplayskip}{2pt} 
        \setlength{\belowdisplayskip}{2pt}
        \[
        \textstyle
        \hat{y}_o^{h, j} \sim \Delta_{\text{vocab}} \bigl( \sum_{t=1}^{T} \alpha_t \cdot \text{softmax}( h_\theta^t ( x_o^h, \hat{y}_o^{h,1:j-1} ) ) \bigr),
        \]
        where $\Delta^{\text{vocab}}$ denotes the simplex on the vocabulary 
    \ENDFOR
\ENDFOR
\STATE \textbf{Train} weak-to-strong model \( f_\phi(\cdot) \) on \( \{(x_o^h, \hat{y}_o^h)\}_{o=1}^{O} \)
\end{algorithmic}
\end{algorithm}

\textbf{Incorporating Prior Term.} Vanilla Adaboost for binary classification uses $\alpha_{t} = \log({\frac{1-\epsilon_t}{\epsilon_t}})$ to calculate votes for each classifier. To keep $\alpha_t$ positive they introduce a weak learning condition, $\epsilon_t < 0.5$, which means that weighted error for each classifier should be better than random. However, using the same weak-learning condition and voting mechanism is not feasible for multi-classification tasks. Thus, \citet{multiboost} uses an additional prior $\log(c-1)$ term, where $c$ is the number of classes, in the calculation of the AdaBoost parameter $\alpha_{t}$. This term serves two purposes: (1) It enables the generation of weak models with accuracy above $\frac{1}{c}$ \%, where $\frac{1}{c}$ \% is random selection accuracy. This is crucial for smaller models and challenging tasks that cannot achieve 50\% accuracy. (2) It ensures that $\alpha$ remains positive. Now, for generative LLMs with large vocabularies, $\log(c-1)$ renders $\alpha_t$ nearly identical, making this approach infeasible. Therefore, we introduce a different method for calculating votes, $\alpha_t \leftarrow \log({\frac{1-\epsilon_t}{\epsilon_t}}) + \log(\frac{1}{1-\epsilon_{pre}}-1)$, where $\epsilon_{pre}$ is the pre-trained model error of the chosen LLM. We provide a theoretical proof (Theorem 1 in Section \ref{theorem1}) regarding how the training error bound of the ensemble weak learners reduces exponentially with the addition of new weak learners while using new $\alpha_t$ method. This term is sensible because $\epsilon_{pre}$ represents the pretrained model error and makes our new weak learning condition to be $\epsilon_t < \epsilon_{pre}$, effectively replacing the binary and multi-class AdaBoost's conditions.

\textbf{Weighted Error Calculation.} Following the new weak learner's condition, the strict condition for each weak expert training is now that the weighted model error (calculated by comparing each token of each sample) must be less than the pre-training error, i.e., $\epsilon_{t} < \epsilon_{pre}$. 
The weighted model error $\epsilon_{t}$ is defined as, $\epsilon_{t} = \sum_{i=1}^{m} \sum_{j=1}^{k_{i}} \mathbbm{1} \{h^t_{\theta} ( \bm{x}^{e}_i, \bm{y}_{i}^{e, j-1}) \neq \bm{y}_{i}^{e, j} \} D_t(i,j) < \epsilon_{pre}.$
Here, $\bm{y}_{i}^{e, j-1}$ is the $(j-1)^\text{th}$ ground-truth token in the answer part. The model $h^t_{\theta} ( \bm{x}_i^{e}, \bm{y}_{i}^{e, j-1})$ predicts the next token and compares it with the ground-truth token $\bm{y}_{i}^{j}$ (during training when ground-truth is still available).

\textbf{Weight Update Equation.} Our sample-weight update equation for each token is $D_{t+1}(i,j) \leftarrow \frac{1}{Z_t} D_t(i,j) e^{\alpha_t \mathbbm{1} \{h^{t}_{\theta} ( \bm{x}_i^{e}, \bm{y}_{i}^{e, j-1}) \neq \bm{y}_{i}^{e, j} \}}$ where $Z_t$ is a normalization factor ensuring that the updated weights satisfy $\sum_{i=1}^{m} \sum_{j=1}^{k_{i}} D_{t+1}(i,j) = 1$. The main idea is to adjust the sample weights to emphasize misclassified examples, thereby guiding the sampling process for next weak learner.

\textbf{Combining experts to Generate Pseudo Answers for Hard Questions.}
To combine the outputs of different experts trained during the various EnsemW2S rounds, we scale the probability distribution for each token generated by the model $h^t_{\theta}$ in round $t$ by its corresponding weight $\alpha_t$. Specifically, we multiply $\alpha_t$ by the probability distribution vector of each token. We then aggregate these weighted distributions across all rounds, normalizing the resulting vector to form a new probability distribution for each token. Using this aggregated distribution, we sample the final predicted token. The process is autoregressive, where the $j^\text{th}$ token of the ``pseudo-answer'' is generated as 
\begin{equation}
\resizebox{0.5\linewidth}{!}{
$
    \widehat{\bm{y}}_o^{h,j} \sim \Delta^{\text{vocab}}\left(\sum_{t=1}^T \alpha_t \cdot \operatorname{softmax}\left(h^{t}_{\theta}\left([\bm{x}_o^{h}, \widehat{\bm{y}}_o^{h,1:j-1
}]\right)\right)\right)
$
}
\end{equation}
where $\Delta^{\text{vocab}}$ represents the simplex over the vocabulary.

By combining the outputs of multiple experts, each trained in different EnsemW2S rounds, the ensemble approach leverages diverse perspectives from the weak models. Each expert contributes their learned strengths, and through weighted aggregation, we diminish the influence of models that are less confident or less effective on certain tokens. This helps reduce variance in the generation process, ensuring that errors from individual weak models are mitigated. The result is a more robust pseudo-labeling system that is better aligned with the true distribution of the hard data, often yielding a performance improvement over any single weak model.

\textbf{Pseudo Answer Generation on Multiple-Choice Datasets.} On multiple-choice Q/A datasets, instead of using generated tokens $\widehat{\bm{y}}^{h}$ as pseudo-answers, we can select one of the choices in the MCQ dataset using negative log-likelihood (NLL). Basically, we calculate the NLL between the choices and $\widehat{\bm{y}}^{h}$ and select the choice with the lowest NLL. For datasets without multiple choices, we can directly use $\widehat{\bm{y}}^{h}$.

\textbf{How does EnsemW2S improve generalization?}
First, Theorem 1 (\ref{theorem1}) shows that the training error decreases as more weak learners are added. Next, the ``Margin theory for generalization'' demonstrates that the margin distribution over the training data improves with each additional weak learner. Consequently, adding new models in the EnsemW2S ensemble reduces training error \emph{without} overfitting, thereby improving generalization.

\textbf{Theorem 1 (Exponential decay of training error with additional weak learners)}\label{theorem1}
\textit{Under the condition \( \epsilon_t < \epsilon_{\text{prev}} \), our token-level ensembling method, EnsemW2S, reduces the training error exponentially:}
\textit{\[
\text{Error}_{\text{train}} = \frac{1}{n} \sum_{i=1}^{m} \sum_{j=1}^{k_i} \sum_{t=1}^{T} \alpha_t \mathbbm{1}\left( h_{\theta}^t(\bm{x}^{e}_i, \bm{y}_{i}^{e, j-1}) \neq \bm{y}_{i}^{e, j} \right) < \prod_{t=1}^{T} Z_t,
\]}
\textit{where \( Z_t < 1 \), \( T \) is the number of EnsemW2S rounds, \( n = \sum_{i=1}^{m} k_i \), and \( k_i \) is the token length of the \( i^{\text{th}} \) easy example \( S^e = \{(\bm{x}^{e}_i, \bm{y}^{e}_i)\}_{i=1}^m \).} We assume tokens are i.i.d. for this proof. See Appendix \ref{appx:theorem1} for full details.

\textbf{Margin Theory for Generalization.} Drawing on results from AdaBoost (binary classification) and SAMME (multiclass boosting), we observe that the margin over training data increases with the number of weak learners, making the ensemble's predictions more confident. Prior work shows increased margins improve generalization. Since we assume i.i.d. tokens, these conclusions can be directly apply. See Appendix \ref{appx:generalization} for details.

\textbf{Train W2S Model.} The strong student model, $f_\phi(\cdot)$, is trained using pseudo answers generated for the hard data $\{(\bm{x}_o^h,\widehat{\bm{y}}_o^h)\}_{o=1}^O$.
Unlike \citet{openAIw2s}, we also include easy human-level data in the training, as it is a sensible choice. For completeness and proper comparison, we additionally run experiments that exclude the easy data, following \citet{openAIw2s}.

\textbf{Ablation Studies.} Logit combination instead of probabilities showed no improvement (see Appendix Figure \ref{fig:plots_prob_logit}). We conducted ablation studies where, instead of treating each token as independent, we used a sliding window of length $L$ while calculating weights and aggregating errors (see Appendix Figure \ref{fig:plots_token_window_70} and \ref{fig:plots_token_window_410}). Different window lengths did not cause significant changes in values, so we ultimately chose a window of  $L=1$. We also explored treating each sample as independent instead of each token as independent in the sample-answer part, finding better results with the latter. This is reasonable since the error calculated using independent-sample weights is less accurate.

\textbf{Evaluation Metric.} We used two metrics to evaluate this Q/A dataset. One is \textbf{(1) Token-wise comparison}, where we compare each predicted token and average the total error, and \textbf{(2) Option-wise comparison}, where we compare the negative log-likelihood (NLL) of the correct answer completion with the NLLs of the incorrect answer completions. Accuracy represents the number of entries where the correct answer completion has the lowest NLL among all choices. We use option accuracy to reported results in the main paper.

\textbf{Computational cost of EnsemW2S.} The primary increase in computational cost compared to the baseline---where a single weak expert supervises the student---comes from multi-LLM decoding. However, its important to consider that unlike other multi-LLM works \cite{anonymous2025collab, du2023improving}, we do not use an additional reward or judge model for combining multi-LLM outputs, significantly reducing hardware demands. Moreover, we rely on the same labeled data rather than collecting new data to train separate models. A detailed computational analysis of each EnsemW2S component is provided in Appendix~\ref{appx:cost}.


%% file: sections/RelatedWorks.tex
\section{Related Work}\label{sec:related}
\textbf{Weak-to-Strong.} \citep{openAIw2s} first introduced w2s generalization for super-alignment, aiming to elicit strong model capabilities using only single weak model supervision. \citep{quant_w2s} provides a theoretical framework for the same with insights on how much w2s improvement can occur, though their work is limited to a few layers of neural networks. \citep{theory_w2s} establishes expansion bounds for w2s generalization under finite data distributions, but focuses solely on binary classification. \citep{zhang2024transcendence} shows that transcendence---surpassing the capability of the training data source---is possible via low-temperature sampling, offering insights relevant to w2s.

Several works have attempted to solve w2s generalization in LLMs. \citep{adaw2s} explores ensemble learning for and scalable oversight for binary classification NLP tasks, but sees limited gains.
\citep{aligner} introduces a model that enhances the alignment of LLMs with human intentions by correcting the residual differences between aligned and unaligned answers by training on a query-answer correction dataset. It enhances w2s generalization by leveraging smaller models for supervisory signals. \citep{sun2024easy} proposes a scalable e2h generalization method, training reward models on simpler tasks and using them to evaluate harder ones. \citep{liu2024co} adapts the hierarchical mixture of experts, using multiple specialized weak supervisors instead of a single generalist for w2s. \citep{weak_weaker} compares LLM training on weak (cheap) vs. strong (expensive) model-generated data and finds that larger weak-model datasets yield better w2s.

\textbf{Ensemble Learning.} Binary Classification Boosting \citep{boost} and multi-classification boosting \citep{multiboost} are common ensemble learning algorithms.
In \citep{juries}, they use a voting mechanism to combine multiple small LLMs instead of a single large LLM to evaluate another LLM and show it performs better than large LLMs. An extended related work section is present in Appendix \ref{app:r_works}.

%% file: sections/4-Experiment_Setup.tex
\section{Experimental Setup}

\paragraph{ID and OOD Data Setup.} To assess our method's improvement in generalization on both in-distribution (ID) and out-of-distribution (OOD) datasets, we use factual question-answering datasets such as ARC \citep{arc} and Quartz \citep{quartz}, as well as more complex mathematical datasets like math-mc \citep{tan2024teaching} and MMLU (categories: high-school mathematics, elementary mathematics, college mathematics) \citep{hendryckstest2021}. All these datasets feature multiple-choice question-answering tasks, facilitating pseudo-label generation based on options.

\textbf{For studying ID performance}, we employ ARC, Quartz, and math-mc datasets, randomly splitting each dataset into two halves. One half is used for training weak experts. We treat the second half as unlabeled, generating pseudo-labels from the trained weak experts. Subsequently, we train the strong student model using both this pseudo-labeled data and the initially labeled half. \textbf{For assessing OOD performance}, we adopt two distinct strategies. First, we partition datasets into easy and hard subsets, training weak experts on labeled easy data, and the strong student on unlabeled hard data combined with labeled easy data (similar to the ID setup). We evaluate performance on a small, labeled subset sharing the hard data distribution. Difficulty levels in math-mc are directly available through question-difficulty ratings, whereas for ARC and Quartz, we generate difficulty ratings via a cross-validation approach described in Appendix~\ref{appx:cross_validation}. Difficulty rating plots for ARC and Quartz are provided in Figures~\ref{fig:diff_arc} and~\ref{fig:diff_quartz}, respectively. The second strategy involves training weak and strong models on one dataset (math-mc) and evaluating performance on a different dataset (MMLU's mathematics categories) to study how cross-data OOD generalization changes.  

\paragraph{Expert and Student Models.} We employ Qwen2.5-1.5B and Qwen2.5-3B models as weak experts, and Qwen2.5-7B as the strong student to study both ID and OOD generalization. Additionally, we explore scaling laws through multiple combinations of weak and strong model pairs using the Quartz dataset. For this scaling analysis, we utilize the Pythia model series, ranging from 70M to 2.8B parameters since pythia even though relatively old model series provided us with a wide range of models sizes. To \textbf{determine the optimal number of AdaBoost rounds}---i.e., how many models to combine---we use a validation set matching the training distribution. We limit our experiments to a maximum of 5 AdaBoost rounds/models.
\paragraph{Baseline.} We compare EnsemW2S to a single weak expert and a strong student trained on outputs of single weak expert. Additional baselines include training the strong student on weak experts' labeled data and strong student's pre-training performance.

%% file: sections/5-Results.tex
\section{Results}

\subsection{Enhancing ID and OOD Generalization of Weak Experts using EnsemW2S}

\begin{figure}[htp]
 \centering
  \includegraphics[trim={0 0 0 10},clip,width=1\linewidth]{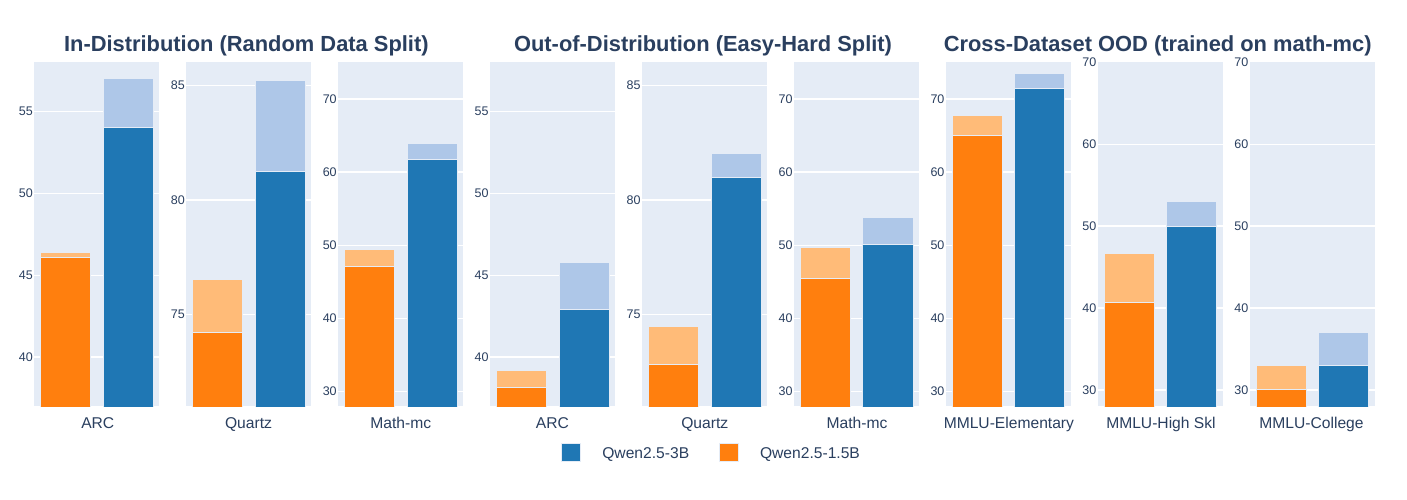}
  \vspace{-2.1em}
\caption{Performance comparison of weak models (single vs. ensemble). Blue bars: Qwen2.5-3B; orange: Qwen2.5-1.5B. Dark shades show single-model performance; light shades show ensemble gains. The bars are grouped into: (1) ID evaluation (first three subplots), where train and test distributions are similar; (2) OOD via easy-hard splits (middle three), where models are trained on the easy and tested on the hard; and (3) cross-dataset OOD generalization (last three), where models trained on the easy of Math-MC are evaluated on three difficulty levels of MMLU.}
\label{fig:weak_expert}
\end{figure}

As depicted in Figure~\ref{fig:weak_expert}, we analyze the in-distribution (ID) and out-of-distribution (OOD) generalization performance of our EnsemW2S-based combination of weak experts for two models, Qwen2.5-1.5B and Qwen2.5-3B. We compare the performance of a single weak model (darker color) against an ensemble of weak models trained on the same labeled dataset, with lighter hues indicating performance improvements. For \textbf{ID generalization}, we observe absolute improvements between of \textbf{0.4-3.02\%, 2.3-4\%, and 2.24-2.26\%} on the ARC, Quartz, and Math-MC datasets, respectively. For \textbf{OOD generalization} based on an \textbf{easy-to-hard data split}, the ensemble achieves improvements between \textbf{1.02-2.84\%, 1.03-1.66\%, and 3.7-4.34\%} on the same datasets. We also evaluate \textbf{cross-dataset OOD performance} by training on the easy split of Math-MC and testing across three difficulty levels in MMLU’s mathematics categories. In this setting, we observe gains between of \textbf{2.07-2.7\%, 3-6\%, and 3-4\%} for elementary, high school, and college-level math questions, respectively. These results suggest that generalization performance is more sensitive to question difficulty than to dataset variation.

As expected, performance of ID data is higher than to OOD data. However, the consistent gains achieved by our ensemble method---despite using the same labeled data as the single-model baseline---highlight its robustness and effectiveness. A few simple baselines are presented in Appendix~\ref{appx:weak_model}.

\subsection{Stronger Expert further Improve Student and its W2S generalization performance.}

\begin{table}[H]
\resizebox{\textwidth}{!}{%
\begin{tabular}{llllllll}
\hline
 &
  \multicolumn{7}{l}{In-Distribution Performance} \\
 &
  \multicolumn{2}{l|}{Qwen2.5-3B} &
  \multicolumn{5}{l}{Qwen2.5-7B} \\ \hline
\multicolumn{1}{l|}{Dataset} &
  \multicolumn{1}{l|}{Single Expert} &
  \multicolumn{1}{l|}{EnsemW2S Expert} &
  \multicolumn{1}{l|}{Student  (Baseline)} &
  \multicolumn{1}{l|}{Student (Ours)} &
  \multicolumn{1}{l|}{Oracle} &
  \multicolumn{1}{l|}{Student on Expert’s data} &
  Pre-trained Student \\ \hline
\multicolumn{1}{l|}{ARC} &
  \multicolumn{1}{l|}{54.02} &
  \multicolumn{1}{l|}{\textbf{57.04}} &
  \multicolumn{1}{l|}{56.74} &
  \multicolumn{1}{l|}{\textbf{58.70}} &
  \multicolumn{1}{l|}{59.39} &
  \multicolumn{1}{l|}{56.04} &
  45.48 \\
\multicolumn{1}{l|}{Quartz} &
  \multicolumn{1}{l|}{81.25} &
  \multicolumn{1}{l|}{\textbf{85.20}} &
  \multicolumn{1}{l|}{84.95} &
  \multicolumn{1}{l|}{\textbf{88.10}} &
  \multicolumn{1}{l|}{88.78} &
  \multicolumn{1}{l|}{83.47} &
  60.00 \\
\multicolumn{1}{l|}{Math-mc} &
  \multicolumn{1}{l|}{61.74} &
  \multicolumn{1}{l|}{\textbf{64.00}} &
  \multicolumn{1}{l|}{62.90} &
  \multicolumn{1}{l|}{\textbf{65.00}} &
  \multicolumn{1}{l|}{65.87} &
  \multicolumn{1}{l|}{62.90} &
  24.73 \\ \hline
 &
  \multicolumn{6}{l}{Out-of-Distribution Performance} &
   \\
 &
  \multicolumn{2}{l|}{Qwen2.5-3B} &
  \multicolumn{5}{l}{Qwen2.5-7B} \\ \hline
\multicolumn{1}{l|}{} &
  \multicolumn{1}{l|}{Single Expert} &
  \multicolumn{1}{l|}{EnsemW2S Expert} &
  \multicolumn{1}{l|}{Student (Baseline)} &
  \multicolumn{1}{l|}{Student (Ours)} &
  \multicolumn{1}{l|}{Oracle} &
  \multicolumn{1}{l|}{Student on Expert’s data} &
  Pre-trained Student \\ \hline
\multicolumn{1}{l|}{ARC} &
  \multicolumn{1}{l|}{43.00} &
  \multicolumn{1}{l|}{\textbf{45.76}} &
  \multicolumn{1}{l|}{45.39} &
  \multicolumn{1}{l|}{\textbf{46.67}} &
  \multicolumn{1}{l|}{51.19} &
  \multicolumn{1}{l|}{45.42} &
  34.00 \\
\multicolumn{1}{l|}{Quartz} &
  \multicolumn{1}{l|}{80.99} &
  \multicolumn{1}{l|}{\textbf{82.02}} &
  \multicolumn{1}{l|}{83.29} &
  \multicolumn{1}{l|}{\textbf{85.71}} &
  \multicolumn{1}{l|}{87.5} &
  \multicolumn{1}{l|}{82.48} &
  56.12 \\
\multicolumn{1}{l|}{Math-mc} &
  \multicolumn{1}{l|}{33.00} &
  \multicolumn{1}{l|}{\textbf{37.00}} &
  \multicolumn{1}{l|}{55.86} & 
  \multicolumn{1}{l|}{\textbf{59.74}} & 
  \multicolumn{1}{l|}{60.91} &
  \multicolumn{1}{l|}{49.00} &
  26.72 \\
\multicolumn{1}{l|}{MMLU-Elementary} &
  \multicolumn{1}{l|}{71.43} &
  \multicolumn{1}{l|}{\textbf{73.28}} &
  \multicolumn{1}{l|}{73.91} &
  \multicolumn{1}{l|}{\textbf{76.19}} &
  \multicolumn{1}{l|}{75.66} &
  \multicolumn{1}{l|}{73.57} &
  44.71 \\
\multicolumn{1}{l|}{MMLU-High-Skl} &
  \multicolumn{1}{l|}{50.00} &
  \multicolumn{1}{l|}{\textbf{53.00}} &
  \multicolumn{1}{l|}{52.59} &
  \multicolumn{1}{l|}{\textbf{53.70}} &
  \multicolumn{1}{l|}{54.00} & 
  \multicolumn{1}{l|}{52.18} &
  23.70 \\
\multicolumn{1}{l|}{MMLU-College} &
  \multicolumn{1}{l|}{33.00} &
  \multicolumn{1}{l|}{\textbf{37.00}} &
  \multicolumn{1}{l|}{46.00} &
  \multicolumn{1}{l|}{\textbf{47.00}} &
  \multicolumn{1}{l|}{47.50} &
  \multicolumn{1}{l|}{44.00} &
  33.00 \\ \hline
\end{tabular}%
}
\caption{The first two columns show performance of a single weak expert and the EnsemW2S based ensemble. Its followed by strong students trained on the single and combined experts. The table also includes an oracle upper bound and two baselines: one trained on weak labels and one showing pretrained performance.}
\label{tab:results_table}
\end{table}

As shown in Table~\ref{tab:results_table}, we study how an ensemble of weak models improves the generalization performance of the strong model. We focus on Qwen2.5-7B for this analysis and evaluate both ID and OOD generalization, following the same setup as before. For \textbf{ID generalization}, we observe \textbf{absolute performance improvement of 2\%, 3.2\%, and 2.1\%} on the ARC, Quartz, and Math-MC datasets, respectively. For \textbf{OOD generalization based on an easy-to-hard} split, we observe gains of \textbf{1.28\%, 2.42\%, and 1.17\%} on the same datasets. We also evaluate \textbf{cross-dataset OOD performance} by training the ensemble on the easy subset of Math-MC and testing on varying difficulty levels within MMLU’s mathematics categories. Here, we observe improvements of \textbf{2.28\%, 1.11\%, and 1\%} for high-school, elementary, and college-level math questions, respectively.

A few notable conclusions: the EnsemW2S-based ensemble of weak experts achieves performance comparable to the W2S-trained strong student, demonstrating the effectiveness of ensembling. It even outperforms a strong model trained on weak experts’ labeled data, despite the total combined model size being comparable in most cases. Moreover, the strong model trained with weak supervision from the ensemble surpasses the baseline which use a single expert model.

\subsection{Improvement in PGR and W2S generalization for all range of model sizes.}

\begin{figure*}[htp]
    \centering
    \begin{minipage}[t]{\textwidth}
        \includegraphics[trim={0 0 0 0},clip,width=1\linewidth]{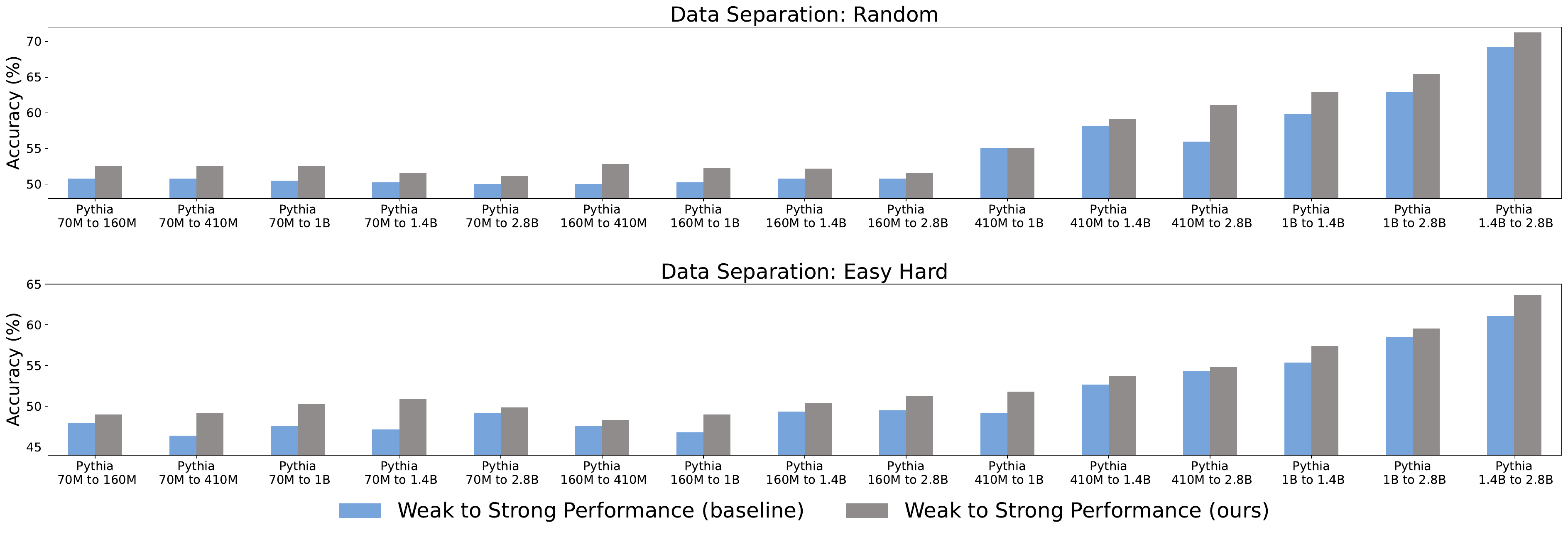}
    \end{minipage}
    \hfill
    \begin{minipage}[t]{\textwidth}
        \includegraphics[trim={0 8cm 0 8cm},clip,width=\linewidth]{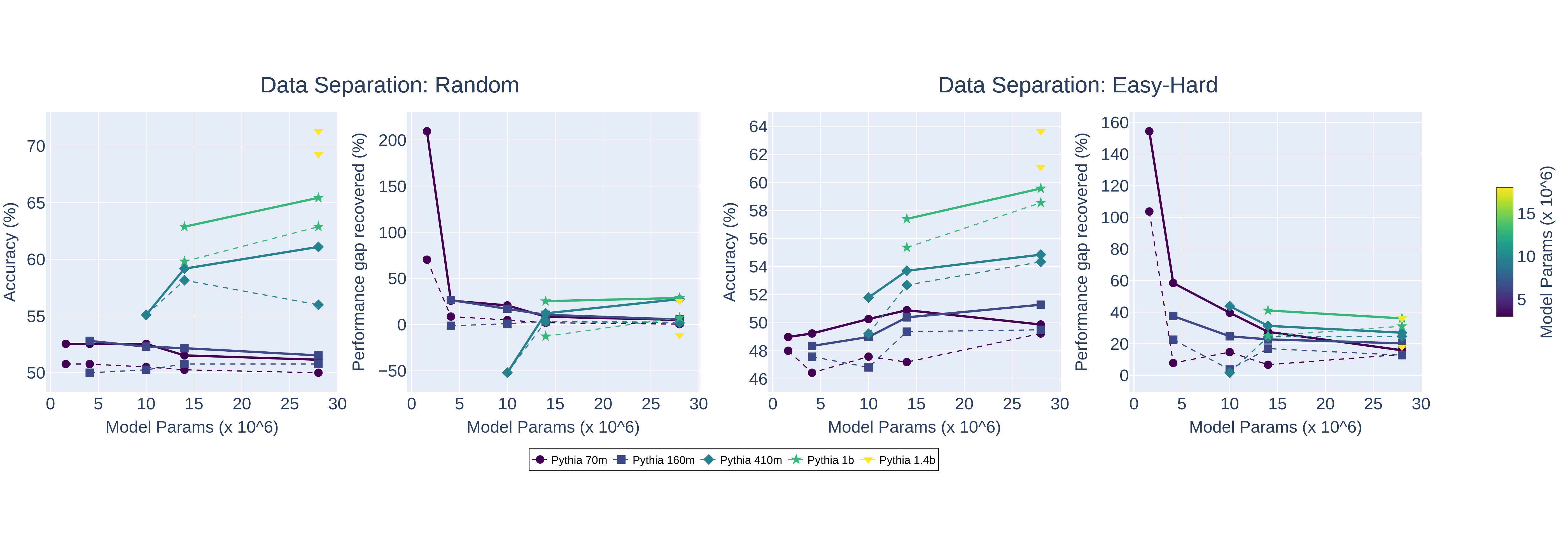}
    \end{minipage}
    \vspace{-16pt}
    \caption{\textbf{(Quartz Data.)} Top: Bar plots comparing W2S generalization of the strong student using our method (gray) vs. baseline (blue) across weak–strong model pairs for SFT on Q/A data—ID (top bars) and OOD (bottom bars). Bottom: Line plots showing accuracy and PGR. Left: ID (random split); Right: OOD (easy–hard split) for E2H generalization.}
    \label{fig:bar_2_quartz_easy}
\end{figure*}

Following \citet{openAIw2s}, we study scaling laws for W2S generalization in Fig.\ref{fig:bar_2_quartz_easy} by creating multiple weak–strong model pairs from the same model family. We use the Pythia model series, which offers a wide range of model sizes. To enable a direct comparison with \citet{openAIw2s}, we train the strong model using only hard data. Results for the Quartz dataset are shown here, while results for ARC are included in Appendix\ref{appx:arc_results}. We also conduct experiments with a binary classification LLM on the SciQ task using AdaBoost as the ensemble method, following the exact experimental setup and code provided by \citet{openAIw2s} (Results present in Appendix~\ref{appx:binary_results}). For all above experiments, we observe consistent improvements—indicating stronger W2S generalization—except in cases where models are too small to learn meaningful information. In the binary setting, we even observe instances of complete W2S generalization, demonstrating the effectiveness of carefully constructed weak experts and highlighting the overall merit of the ensemble approach.

%% file: sections/6-Conclusion.tex
\section{Conclusion}\label{sec:conclusions}
This paper presents a method to iteratively train weak experts by exposing them only to human-level labeled data, enabling them to better generalize to superhuman-level data. These weak supervisors are then combined at the token level to generate stronger supervision, improving W2S generalization for the strong student model. Our method shows significant improvement on both in-distribution and out-of-distribution datasets.

%% file: sections/Appendix.tex
\section{Limitation and Future Work}
This work only explores the SFT phase. While SFT is an important part of the LLM learning pipeline, our future work will focus on developing weak supervision in the reward modeling phase. Another interesting future direction would be to improve the combination of tokens in the decoding phase by replacing the classical AdaBoost algorithm with more adaptive ensemble learning methods. We hope this work sparks discussion on combining multiple LLMs to improve w2s generalization.

\textit{Computational Overhead:} For fully generative tasks, multiple forward passes are required in an autoregressive manner. At each step, the final voted token is input to all LLMs to predict the next token. This increases generation time, which can be mitigated using efficient decoding algorithms like speculative decoding. Addressing this also forms part of our future work. \textit{Smaller Models:} Another limitation is of all w2s work is they attempt to mimic the weak and strong setting as an analogy to the realistic problem and cannot test on a real human with super-human model.

\section{Broader Impact}
The proposed framework for weak-to-strong (w2s) generalization using ensembles of weak language models (LLMs) has significant implications across various domains. By demonstrating that multiple weak supervisors can effectively train more powerful models, our research addresses the critical challenge of superalignment, potentially transforming how advanced AI systems are developed and supervised. This approach could democratize access to powerful AI technologies by reducing reliance on scarce, high-quality labeled data and enabling more inclusive participation in AI development. Furthermore, our method encourages the creation of robust AI systems capable of tackling complex problems, which can drive advancements in fields such as healthcare, education, and scientific research. However, careful consideration must be given to ethical implications, ensuring that the deployment of these advanced models aligns with societal values and mitigates risks associated with misuse or unintended consequences.

\section{Related Works}\label{app:r_works}

\paragraph{Weak-to-Strong Generalisation:} (Continue from the main manuscript) \citet{guo2024vision} introduces an dynamic adjustable loss function for weak-to-strong supervision. \citet{hase2024unreasonable} demonstrates that current language models can achieve high performance on difficult tasks by training on simpler, cleanly labeled data, thus avoiding the high costs and noise associated with hard data labeling. None of these works focused on making the weak teachers, less weak but only focus on improving transfer learning and correction of weak labels. Thus, our method can be combined with all ideas focused on improving transfer learning.

\paragraph{Multi-LLM learning:} There are numerous works involving the collaboration of multiple LLMs. \cite{multi_LLM4} proposes Reinforcement Learning with Guided Feedback (RLGF), where a dynamic black-box guide like GPT-3 is used to fine-tune large language models. \cite{multi_LLM5} introduces Direct Nash Optimization (DNO), a scalable algorithm that combines contrastive learning with general preference optimization. \cite{Multi_LLM} presents MEDUSA, an innovative framework designed to accelerate inference in large language models by introducing multiple decoding heads, enabling simultaneous prediction of several tokens, and enhancing efficiency through reduced decoding steps and parallel processing capabilities. \cite{Multi_LLM2_col} proposes Co-LLM, a collaborative decoding framework that interleaves token-level generations from multiple models. This method optimizes the latent variable model for marginal likelihood, allowing a base model to decide when to generate tokens itself or utilize an assistant model, thereby improving performance across various specialized tasks without direct supervision. \cite{Multi_LLM3} introduces a novel collaborative decoding framework aimed at improving the factuality of large language models by employing a critical token classifier. This approach strategically uses both pre-trained and aligned models to selectively generate critical tokens, significantly enhancing the model's ability to maintain factual accuracy without compromising the diversity of the generated content.

Additionally, \cite{controlled_decoding} introduces Controlled Decoding (CD), a method for aligning language model outputs with desired outcomes using a separate prefix scorer module. This approach allows multi-objective RL without additional training and performs well on benchmarks, bridging the gap between token-level control and sequence-level best-of sampling strategies.

\section{Adaboost for combing weak Binary classification LLM.}
\label{appx:adaboost}

AdaBoost is an ensemble learning algorithm that combines multiple weak classifiers, such as decision stumps, to create a strong classifier. It works iteratively by focusing on the samples that are hardest to classify, assigning them higher weights in each subsequent iteration. Weak classifiers are trained one at a time, and their contributions are weighted based on their accuracy. The final prediction is made by taking a weighted majority vote of all weak classifiers. AdaBoost is known for its ability to improve generalization by focusing on difficult cases and is often resistant to overfitting with simple weak learners. However, it can struggle with noisy data if overemphasis is placed on misclassified samples. Its also presented as Algorithm \ref{alg:adaboost}.

We use Adaboost to conduct a thought experiment to test the ensemble idea in w2s generalization for simple classification task. This is also the first task evaluated by \citet{openAIw2s}. We use vanilla AdaBoost (Algorithm \ref{alg:adaboost}) to generate answers to hard questions, $\bm{x}^h$, from each weak LLM teacher, $ h_{\theta}^t (\bm{x}^h)$ for $t\in \{1,\ldots,T\}$, where $T$ is max Adaboost round. AdaBoost iteratively reweights training examples, and trains weak teachers one at a time focusing on hard-to-classify samples, as described in Line 5 of Algorithm \ref{alg:adaboost}. The only requirement is they perform better than random, thus satisfying the weak learning condition. A weighted majority vote aggregates their outputs to produce a pseudo-label, $\mathbbm{1}(\sum_{t=1}^T \alpha_t h_{\theta}^t (\bm{x}^h)>0)\in \{0,1\}$, where ${\alpha_t}$ are hyperparameters weighing each weak teacher based on accuracy. A detailed mathematical summary is provided below.

Let the training dataset consist of $m$ samples:
\[
\{(x_i, y_i) \mid i = 1, 2, \dots, m\}, \quad x_i \in \mathbb{R}^d, \quad y_i \in \{-1, +1\}.
\]

Each weak learner $h_t(x)$ outputs a prediction $h_t(x_i) \in \{-1, +1\}$. The goal is to sequentially train weak learners such that the combined model minimizes the classification error. A weight distribution $D_t(i)$ is maintained over the training samples at each iteration $t$, where:
\[ D_t(i) \geq 0, \quad \sum_{i=1}^m D_t(i) = 1. \]
Initially, all samples are equally weighted: $D_1(i) = \frac{1}{m}, \quad \forall i$

\paragraph{Training the Weak Learners:}
For each iteration $t = 1, 2, \dots, T$,  train a weak learner $h_t(x)$ using the current weight distribution $D_t$. Compute the weighted error: \[\epsilon_t = \sum_{i=1}^m D_t(i) \cdot \mathbb{I}(h_t(x_i) \neq y_i),\] where $\mathbb{I}(\cdot)$ is the indicator function.

\paragraph{Weak Learner Weight}
Assign a weight $\alpha_t$ to the weak learner based on its performance:
\[ \alpha_t = \frac{1}{2} \ln \left( \frac{1 - \epsilon_t}{\epsilon_t} \right) \] Intuition behind it is that if $\epsilon_t$ is small, $\alpha_t$ is large, giving more importance to the weak learner. If $\epsilon_t = 0.5$, $\alpha_t = 0$, indicating no contribution to the ensemble. $\epsilon_t > 0.5$ is undesirable, as the weak learner performs worse than random guessing.

\textbf{Update the weights} of the training samples to focus on misclassified samples:
\[
D_{t+1}(i) = \frac{D_t(i) \exp(-\alpha_t y_i h_t(x_i))}{Z_t},
\]
where $Z_t$ is a normalization factor ensuring $\sum_{i=1}^m D_{t+1}(i) = 1$:
\[
Z_t = \sum_{i=1}^N D_t(i) \exp(-\alpha_t y_i h_t(x_i)).
\]

Misclassified samples ($y_i \neq h_t(x_i)$) receive higher weights, making them more influential in the next iteration. The \textbf{final strong classifier} $H(x)$ is a weighted majority vote of the weak learners:
\[
H(x) = \operatorname{sign} \left( \sum_{t=1}^T \alpha_t h_t(x) \right).
\]

\paragraph{Generalization Abilities:} AdaBoost improves generalization by maximizing the margins on the training set. The margin for a sample $(x_i, y_i)$ is defined as:
\[
\text{Margin}(x_i) = y_i \sum_{t=1}^T \alpha_t h_t(x_i).
\]
AdaBoost aims to increase the margin for all samples, reducing the chance of misclassification.







\paragraph{Summary of Key Properties}
\begin{enumerate}
    \item \textbf{Sequential Training:} Weak learners are trained iteratively, with weights updated to focus on difficult samples.
    \item \textbf{Weighting Scheme:} Misclassified samples are emphasized in subsequent iterations.
    \item \textbf{Generalization:} AdaBoost achieves strong generalization by maximizing margins and minimizing exponential loss.
    \item \textbf{Flexibility:} It can work with any weak learner as long as the learner achieves performance slightly better than random guessing.
\end{enumerate}

\begin{algorithm}[!htbp]
\caption{AdaBoost  \cite{boost}}
\label{alg:adaboost}
\begin{algorithmic}
\REQUIRE Training Dataset \(S = \{(x_i, y_i)\}_{i=1}^m \sim D^m\)
\STATE $T =$ AdaBoost iterations 
\STATE \(\vec{D}_1(i) \leftarrow \frac{1}{m} \forall i \in [m]\) 
\FOR{\(t \leftarrow 1\) to \(T\) } 
    \STATE $h_t$ such that $\epsilon_{t} = \sum_{i=0}^{m} \mathbbm{1} \{h_t ( x_i ) \neq y_i \} \vec{D}_t(i) < \frac{1}{2} $
    \STATE \(\alpha_t \leftarrow \frac{1}{2}\log{\frac{1-\epsilon_t}{\epsilon_t}}\) 
    \STATE \(Z_t \leftarrow 2\sqrt{\epsilon_t(1-\epsilon_t)}\) 
    \STATE \(\vec{D}_{t+1} \leftarrow \frac{1}{Z_t} \vec{D}_t e^{-\alpha_t y_i h_t(x_i)}\) 
    \STATE \(g \leftarrow \sum_{t=1}^T \alpha_t h_t\) 
\ENDFOR
\STATE Return \(h(x) = \operatorname{sign}(g)\)
\end{algorithmic}
\end{algorithm}

\section{Details on the EnemW2S Methodology}

\subsection{Detailed Flowchart}

\begin{figure}[H]
    \centering
    \includegraphics[trim={0 0 0 0},clip,width=0.9\linewidth]{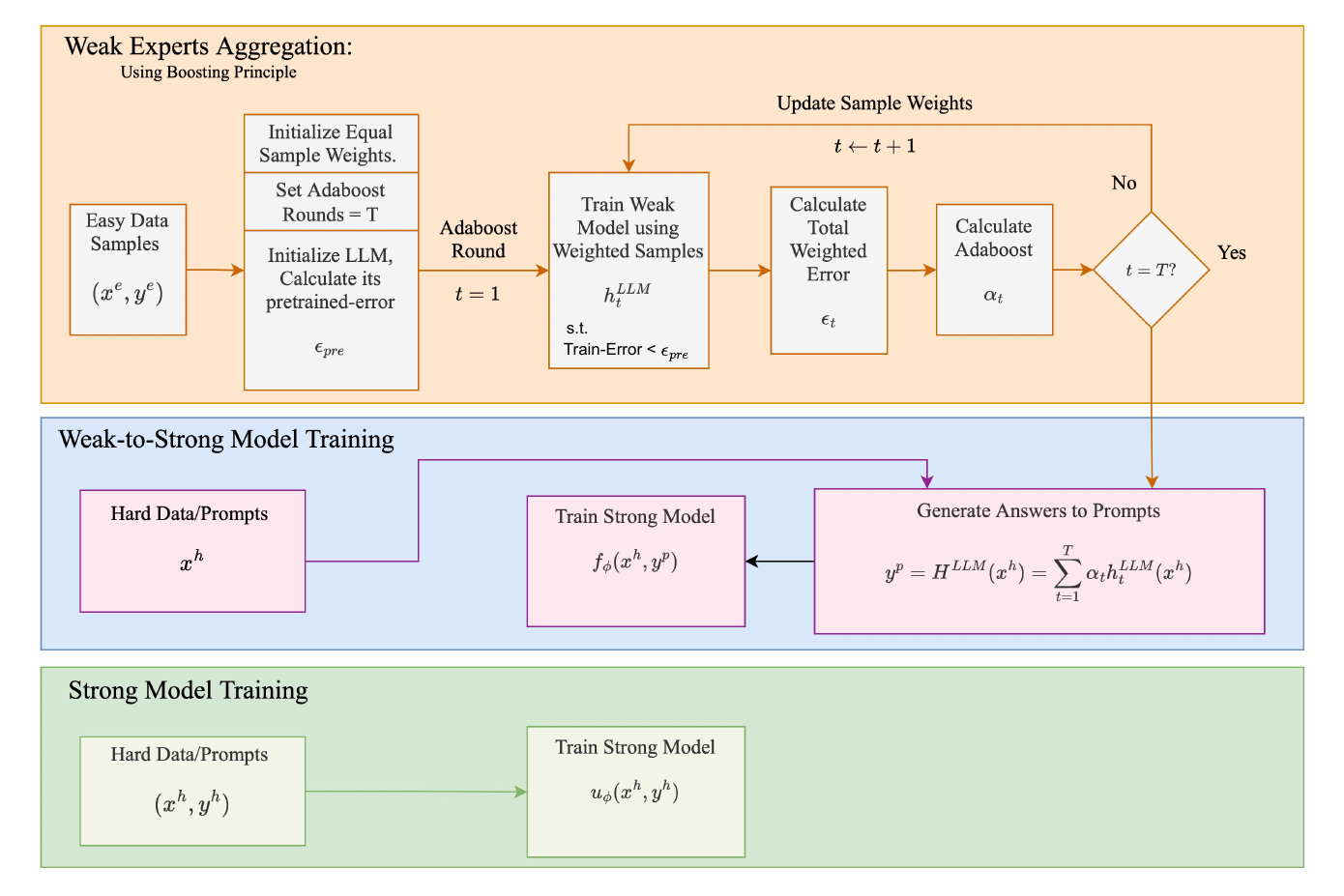}
    \caption{This figure explains our pipeline for easy-to-hard generalization using w2s generalization in complete detail including the algorithm and data flow. We train weak models on easy data and strong models on hard data. A transfer model is trained using pseudo labels generated by the weak model on the hard data. Ultimately, we aim to improve the Performance Gap Recovered (PGR).}
    \label{fig:train_flow_detail}
\end{figure}

\subsection{Important Notations}\label{appx:notions}

Easy Data: $\{(\bm{x}^e_{i},\bm{y}^e_{i})\}_{i=1}^{m}$ 

Hard Data: $\{(\bm{x}^h_{o},\bm{y}^h_{o})\}_{o=1}^{O}$

Total number of Easy Data points: $m$

Total number of Hard Data points: $O$

Total EnsemW2S-AdaBoost Rounds: $T$
Weak Teachers: $\{h^t_{\theta}\}_{t=1}^{T}$

Strong Student (Oracle): $u_{\phi}$

Weak-to-Strong model: $f_{\phi}$

Total number of tokens in the answer part of each sample $i$: $k_{i}$

AdaBoost voting parameter: $\{\alpha_{t}\}_{t=1}^{T}$

EnsemW2S-AdaBoost token-sample weights for $i^{th}$ sample and $j^{th}$ token: $\{D_{t}(i,j)\}_{t=1}^{T}$

Pre-trained Model error: $\epsilon_{pre}$

EnsemW2S-AdaBoost's weighted model error for round $t$: $\epsilon_{t}$

\subsection{Intuition Behind Prior Term in EmsemW2S}
\label{appx:prior_term}

The calculation of $\alpha$ cannot rely solely on error, $\epsilon$, as the traditional Adaboost method is valid only when $\epsilon < 0.5$. Applying the same equation in our context could yield negative $\alpha$ values. We introduce a prior term, $\log(\frac{1}{1-\epsilon_{pre}}-1)$. Below we provide justification and proof regarding exponential decay of the training error using the new prior term. 

Existing works on multi-class classification Adaboost \cite{multiboost} suggest using $\frac{1}{c}$ (where c is the number of classes) in the prior term, $\log(c - 1)$, as $\frac{1}{c}$ represents the random performance of the model. However, when c (the number of classes) becomes very large, the $\log(c - 1)$ term also grows significantly, causing the $\alpha$ parameters of Adaboost to become nearly identical and, consequently, less useful. To address this, we introduce a pre-training error term, $\epsilon_{pre}$, which represents an upper bound on the sample error. We then use $1 - \epsilon_{pre}$ (a lower bound on accuracy) as a replacement for the $\frac{1}{c}$ term, as our model's lowest possible accuracy is $1 - \epsilon_{pre}$, not $\frac{1}{c}$.

\section{[Theorem 1:] Training Error Bound Exponential Decay.}\label{appx:theorem1}

Training Error of the combined weak learner is defined as the following: 
\[\text{Err}(H) = \frac{1}{n}\sum_{i=1}^{m} \sum_{j=1}^{k_i} \sum_{t=1}^{T} \alpha_t \mathbbm{1} (h_{\theta}^t ( \bm{x}^{e}_i, \bm{y}_{i}^{e, j-1}) \neq \bm{y}_{i}^{e, j} ) \tag{1}\]  where $T$ is EmsemW2S rounds, $n = \sum_{i=1}^{m} k_{i}$, $k_{i}$ is the token length in the $i^{th}$ easy example $S^{e} = \{(\bm{x}^{e}_i, \bm{y}^{e}_i)\}_{i=1}^m$.

AdaBoost minimizes an upper bound on the exponential loss rather than directly minimizing the classification error. The exponential loss for our case is:

\[
L = \sum_{i=1}^{m} \sum_{j=i}^{k_i} D_1(i,j) \cdot e^{\sum_{t=1}^{T} \alpha_t \mathbbm{1} (h_{\theta}^t ( \bm{x}^{e}_i, \bm{y}_{i}^{e, j-1}) \neq \bm{y}_{i}^{e, j} )} 
\tag{2}\]

To analyze how the training error decreases, consider the weight update rule:

\[D_{t+1}(i,j) \leftarrow \frac{1}{Z_t} D_t(i,j) e^{\alpha_t \mathbbm{1} (h_{\theta}^t ( \bm{x}^{e}_i, \bm{y}_{i}^{e, j-1}) \neq \bm{y}_{i}^{e, j} ) } \text{ for all } i \in [m], j \in [k_{i}] \tag{3}\]

where $Z_t$ is the normalization factor:

\[
Z_t = \sum_{i=1}^{m} \sum_{j=i}^{k_i} D_t(i,j) \cdot e^{\alpha_t \mathbbm{1} (h_{\theta}^t ( \bm{x}^{e}_i, \bm{y}_{i}^{e, j-1}) \neq \bm{y}_{i}^{e, j} )} 
\tag{4}\]

Now we want the \textbf{cumulative training error} after \( T \) rounds to be bounded by (More details regarding the following equation is in section \ref{appx:z_t}):

\[
\text{Err}(H) \leq \prod_{t=1}^{T} Z_t^{\frac{1}{\epsilon_{pre}}}
\tag{5}
\]

Now for $\text{Error}_{\text{train}}$ to exponentially decay with $t$, $Z_t < 1$. For $Z_t < 1$ we must have $\alpha_t > 0$. 

Thus, we want $$\log{\frac{1-\epsilon_t}{\epsilon_t}} + \log(\frac{1}{1-\epsilon_{pre}}-1) > 0$$

$$(\frac{1-\epsilon_t}{\epsilon_t} )*(\frac{1}{1-\epsilon_{pre}}-1) > 1$$

Solving this we get

$$\epsilon_t < \epsilon_{pre}$$

This forms the our weak learning condition. This basically means that each new weak learner (LLM) that is included in the ensemble should have error less than pretrained error. This error could also be the worst possible error below which any of the LLMs would not go.

\subsection{Calculating Training Error Bound for EnsemW2S}
\label{appx:z_t}

In this section we will derive equation 5 of the previous section.

We know our models weights from previous section:  
\(\alpha_t = \ln\!\bigl(\frac{1-\epsilon_t}{\epsilon_t}\bigr)+\ln(\frac{\epsilon_{pre}}{1-\epsilon_{pre}})\). We can break the equation 4 in the following form,
\[
Z_t=(1-\epsilon_t)+\epsilon_t\,e^{\alpha_t}
      = (1-\epsilon_t)(\frac{1}{1-\epsilon_{pre}}) = \frac{\gamma_t - \epsilon_{pre}}{1-\epsilon_{pre}} .
\]


\vspace{0.4em}
\noindent
\textbf{Potential.}  
Define the exponential potential and multi-class margin $m_{i}$
\[
W_{T+1}
  \;=\;\frac1n \sum_{i=1}^{n}
           \exp\!\bigl(-\lambda m_i^{(T)}\bigr) \tag{6} \]
 \[ \quad
  m_i^{(T)} =
  \frac12\!\Bigl(
    F_{\bm{y}_{i}^{e, j}}^{(T)}(\bm{x}^{e}_i, \bm{y}_{i}^{e, j-1})-
    \max_{k\neq \bm{y}_{i}^{e, j}} F_k^{(T)}(\bm{x}^{e}_i, \bm{y}_{i}^{e, j-1})
  \Bigr) \tag{7}
\]
where \(F_k^{(T)}\) is the accumulated vote after \(T\) rounds and $\lambda = \frac{1}{\epsilon_{pre}}$.

Mathematically, we can therefore write  $F_k^{(T)}$ as 

$$F^{(T)}_k(\bm{x}^{e}_i, \bm{y}_{i}^{e, j-1})=\sum_{t=1}^T \alpha_t \mathbf{1}\left[ h_{\theta}^t ( \bm{x}^{e}_i, \bm{y}_{i}^{e, j-1}) = \bm{y}_{i}^{e, j} \right]$$ 

In a similar way we can define the following final hypothesis,

$$H(\bm{x}^{e}_i, \bm{y}_{i}^{e, j-1})=\arg \max _{k \in[K]} F_k(\bm{x}^{e}_i, \bm{y}_{i}^{e, j-1})$$

A short algebraic check (see Zhu et al., 2006) gives the \emph{telescoping
relation: } $W_{t+1}=W_t Z_t^{(K-1) / K}$. Therefore,
\[
W_{T+1} \;=\; \prod_{t=1}^{T} Z_t^{\,\lambda}.
\tag{8}
\]

\vspace{0.4em}
\noindent
\textbf{Bounding the indicator.} If $H(\bm{x}^{e}_i, \bm{y}_{i}^{e, j-1}) \neq \bm{y}_{i}^{e, j}$, then $m_i \leq 0$. Hence
 \(\mathbbm{1}[H(\bm{x}^{e}_i, \bm{y}_{i}^{e, j-1})\neq \bm{y}_{i}^{e, j}] \le e^{-m_i^{(T)}}\) and
\(e^{-m_i^{(T)}} \le e^{-\lambda m_i^{(T)}}\;\)(\(\lambda<1\)),
\[
\text{Err}(H)
\;=\;\frac1n\sum_{i=1}^n\mathbbm{1}[H(\bm{x}^{e}_i, \bm{y}_{i}^{e, j-1})\neq \bm{y}_{i}^{e, j}]
\;\le\; \frac{1}{n} \sum_{i=1}^n e^{-\lambda m_i}.
\tag{9}
\]

\vspace{0.4em}

The $m_i$ appearing in Eq. (2) is the margin after all $T$
rounds. Therefore the right-hand side of equation 8 is exactly the uniform-average potential in (7):

\[\boxed{\text{Err}(H) 
\;\le\; W_{T+1} = \prod_{t=1}^{T} Z_{t}^\frac{1}{\epsilon_{pre}} }\]

\noindent
\textbf{Final bound.}
Using \(1+z\le e^{z}\;(z\in[0,1])\) we have
\(Z_t \le e^{K\gamma_t}\). Here $K = \frac{1}{1-\epsilon_{pre}}$ and $\gamma_{t} = 1 - \epsilon_{t} - \frac{1}{K}$
\[
\boxed{\;
\text{Err}(H)
\;\le\;
\exp\!\Bigl(
  - (K\!-\!1)\sum_{t=1}^{T}\gamma_t
\Bigr)
\;}
\]
Thus, if every \(\gamma_t>0\) the empirical error decays
\emph{exponentially} with \(T\).

\hfill\(\square\)

\section{Margin theory for generalization}\label{appx:generalization}

For any \(\theta>0\) define the small–margin set  
\(\mathcal{M}(\theta)=\{i:m_i^{(T)}\le\theta\}\).
From the derivation above
\[
\Pr_{i\in S}\bigl[m_i^{(T)}\le\theta\bigr]
\;\le\;
e^{\lambda\theta}\,W_{T+1}
\;\le\;
\exp\!\Bigl(
  \lambda\theta-(K\!-\!1)\sum_{t=1}^{T}\gamma_t
\Bigr).
\tag{B.1}
\]

\paragraph{VC-style margin bound (multi-class).}
Let \(d\!=\!\mathrm{VCdim}(\mathcal{H})\) for the weak-learner
family.  A direct extension of the binary margin bound of
Schapire, Freund, Bartlett \& Lee (1998) to \(K\) classes
(see Schapire \& Singer, 1999; Zhu et al., 2006) states:

\begin{theorem}[Generalisation via Margins]
\label{thm:mc_margin}
For any \(\theta\in(0,1]\) and \(\delta\in(0,1)\), with probability
\(\ge 1-\delta\) over the draw of an \(n\)-sample \(S\),
the boosted classifier \(H\) satisfies
\[
\Pr_{(x,y)\sim\mathcal{D}}\!\bigl[H(\bm{x}^{e}_i, \bm{y}_{i}^{e, j-1})\neq \bm{y}_{i}^{e, j} \bigr]
\;\le\;
\Pr_{i\in S}\!\bigl[m_i^{(T)}\le\theta\bigr]
\;+\;
\sqrt{\frac{c\bigl(d\log n+\log(1/\delta)\bigr)}
           {n\,\lambda^{2}\theta^{2}}}\!,
\]
where \(c\) is an absolute constant.
\end{theorem}

\paragraph{Combining (B.1) with Thm.\,\ref{thm:mc_margin}.}
Using the empirical bound (B.1) inside the theorem yields
\[
\boxed{\;
\Pr_{\mathcal{D}}\!\bigl[H(\bm{x}^{e}_i, \bm{y}_{i}^{e, j-1})\neq \bm{y}_{i}^{e, j}\bigr]
\;\le\;
\exp\!\Bigl(
   \lambda\theta-(K\!-\!1)\sum_{t=1}^{T}\gamma_t
\Bigr)
\;+\;
\sqrt{\frac{c\bigl(d\log n+\log(1/\delta)\bigr)}
           {n\,\lambda^{2}\theta^{2}}}\;.
}
\]

Choose a fixed margin target \(\theta>0\).  
Because the first term decays exponentially in
\(T\) (via \(\sum_t\gamma_t\)) and the second shrinks
as \(O(n^{-1/2})\), the overall test error can be made arbitrarily
small by taking enough boosting rounds and/or sufficiently many training
examples.

\subsection{Discussion}

\begin{itemize}
\item \textbf{Empirical error.}  
      Theorem \ref{appx:z_t} shows that EnsemW2S drives the
      training error to zero at an exponential rate provided every weak
      learner beats random guessing
      \((\epsilon_t<\epsilon_{pre})\).

\item \textbf{Generalization.}  
      Theorem \ref{appx:generalization} connects this fast
      margin growth to a quantitative VC-style bound on test error,
      mirroring the classical binary AdaBoost margin theory.

\item \textbf{Practical implication.}  
      In practice, once the empirical small-margin fraction
      \(\Pr_{i\in S}[m_i\le\theta]\) plateaus, additional boosting offers
      diminishing returns unless more data (larger \(n\)) or a richer
      weak-learner class (larger \(d\)) is available.
\end{itemize}

\section{Experimental Setup}\label{appx:cross_validation}
\subsection{Cross Validation method for difficulty rating generation}
\textbf{Easy $(\bm{x}^e,\bm{y}^e)$ and Hard $(\bm{x}^h,\bm{y}^h)$ Data Split.} We generate difficulty ratings using cross-validation in $n$ folds, where the model trains on $(n-1)$ splits and tests on the remaining split, repeating for all folds. The errors aggregated across the samples serve as difficulty ratings. Low-rated samples are used for weak model training, while high-rated samples are used to generate strong model training data and testing data at random. This cross-validation approach, applied to both classification and generation tasks, ensures robust difficulty splits. Additional details and plots of difficulty ratings are provided in Figures~\ref{fig:diff_sciq}, \ref{fig:diff_arc}, and \ref{fig:diff_quartz} in the Appendix.
Additionally, the continuous difficulty labels provided by \citet{ding2024easy2hardbenchstandardizeddifficultylabels} for math, programming, chess, and reasoning datasets enable flexible easy-hard data splits at any chosen difficulty threshold. As this aspect is independent of our core message, we leave its exploration to future work.

\subsection{Difficulty rating for all the datasets}
We use GPT-2 for binary classification and pythia-160m for SFT task's easy and hard splitting. We use the same training parameters as used in the training of the actual w2s results. 

\begin{figure}[htp]
 \centering
  \includegraphics[trim={0 0 0 0},clip,width=0.5\linewidth]{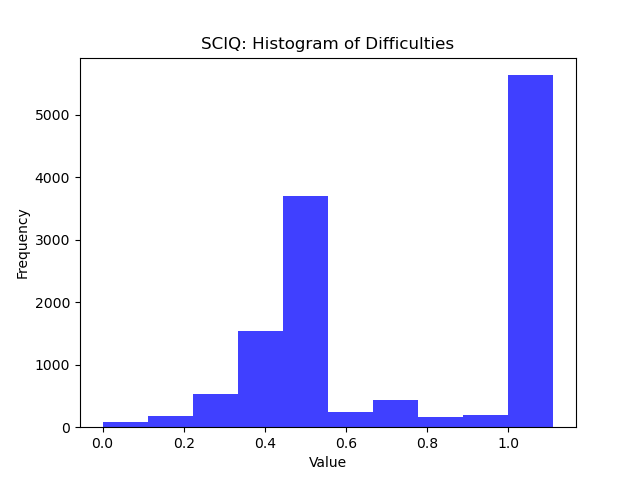}
\caption{This figure shows the difficulty rating distribution of sciq dataset.}
\label{fig:diff_sciq}
\end{figure}

\begin{figure}[htp]
  \centering
  \includegraphics[trim={0 0 0 0},clip,width=0.5\linewidth]{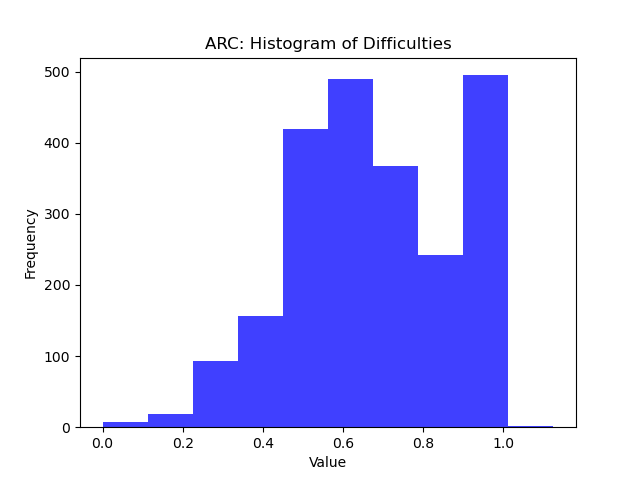}
\caption{This figure shows difficulty rating distribution of ARC dataset.}
\label{fig:diff_arc}
\end{figure}

\begin{figure}[htp]
\centering
  \includegraphics[trim={0 0 0 0},clip,width=0.5\linewidth]{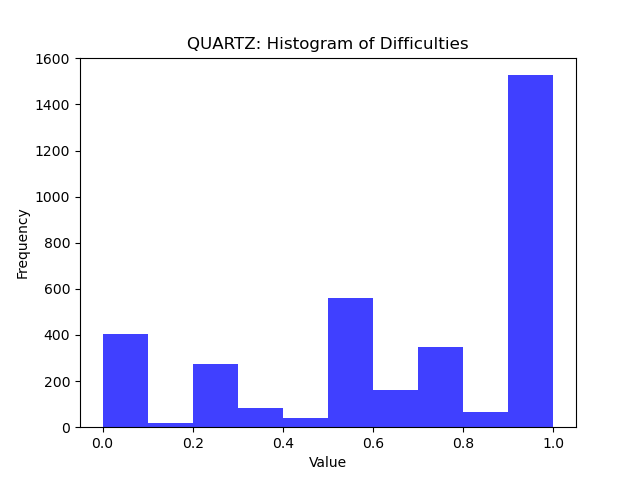}
\caption{This figure shows difficulty rating distribution of quartz dataset.}
\label{fig:diff_quartz}
\end{figure}

\subsection{[Ablation Study: ] Comparison between probability based combination with logit based combination of the tokens, during generation and evaluation of combined weak experts.}

\begin{figure}[H]
\centering
  \includegraphics[trim={0 0 0 0},clip,width=0.5\linewidth]{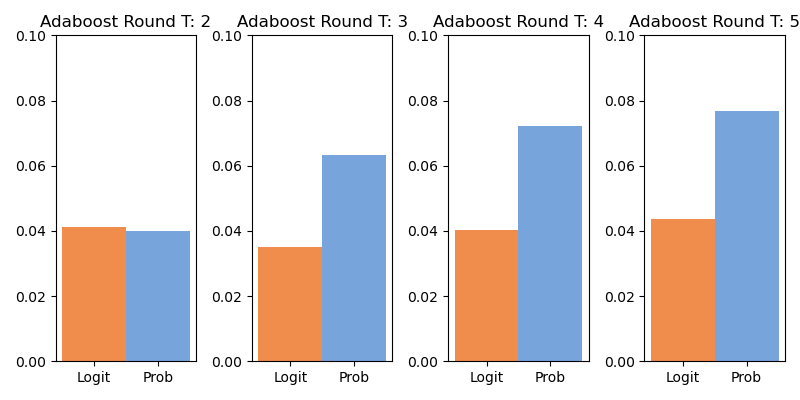}
\caption{This figure compares probability-based combination with logit-based combination of the tokens across different AdaBoost rounds. Here we show improvement from the baseline where baseline is single model. The orange bars represent logit-based combination, while the blue bars represent probability-based combination, showing that probability-based combination performs better. }
\label{fig:plots_prob_logit}
\end{figure}

\subsection{[Ablation Study: ] Comparison between different window lengths for ``sample and token weighing''.}

\begin{figure}[H]
\centering
  \includegraphics[trim={0 0 0 0},clip,width=0.5\linewidth]{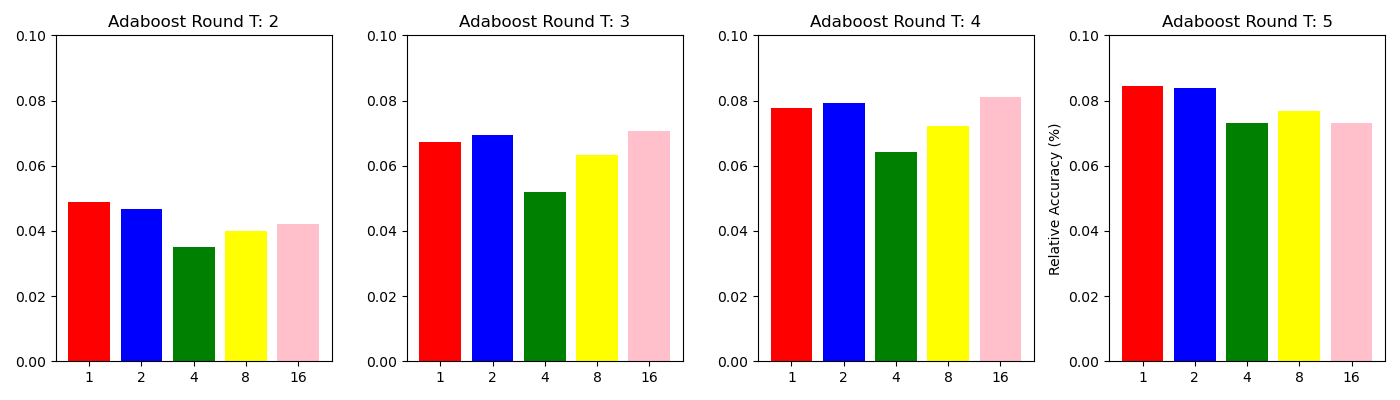}
\caption{This figure compares different token window lengths for the Pythia 70M model across various AdaBoost rounds. The plots show improvements over the baseline, where the baseline represents a single model. The different bars (red, blue, green, yellow, and pink) correspond to window lengths of 1, 2, 4, 8, and 16, respectively. We observe that, overall, all window lengths perform similarly. Window length in EmsemW2S plays a role only during sampling step.}
\label{fig:plots_token_window_70}
\end{figure}

\begin{figure}[H]
\centering
  \includegraphics[trim={0 0 0 0},clip,width=0.5\linewidth]{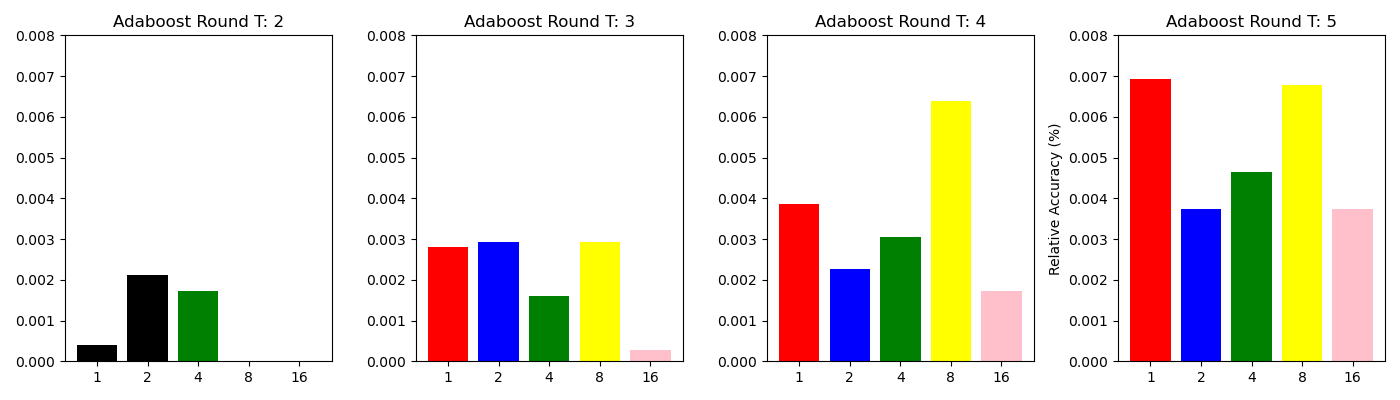}
\caption{This figure compares different token window lengths for the Pythia 410M model across various AdaBoost rounds. The plots show improvements/decline over the baseline, where the baseline represents a single model. Thus, the black colored bars show decline. The different bars (red, blue, green, yellow, and pink) correspond to window lengths of 1, 2, 4, 8, and 16, respectively. We observe that, overall, all window lengths perform similarly. Window length in EmsemW2S plays a role only during sampling step.}
\label{fig:plots_token_window_410}
\end{figure}

\section{Cost Analysis of EmsemW2S}\label{appx:cost}
\paragraph{Training Cost of Weak Learners:} Each weak learner is trained sequentially, as its performance is contingent upon the outputs of the preceding weak learner. Consequently, while the GPU load may be lower because each weak learner is small as compared to the collection of weak-learners, the overall training time is directly proportional to the number of weak learners utilized.

This is because the input and output token count for each weak learner during training remains approximately constant, as suggested by Adaboost. Only the frequency of samples are adjusted based on weights. In EnsemW2S we sample the tokens by token-weights but eventually combine the sampled tokens while masking the ones not sampled, thus keeping the total tokens approximately similar and training time for each weak-learner independent of the tokens sampled.
In the practical superalignment case, pre-trained weak learners can be used, which may further mitigate concerns regarding sequentially training the weak learners.

\paragraph{Strong Model Training and Inference:} The strong model is trained using labels generated by the weak learners and is evaluated on standard datasets. Therefore, the training cost and inference cost associated with the strong model remains unchanged.

\paragraph{Inference Cost of Weak Learners:} The generation process can be executed in parallel as well as sequentially, resulting in a GPU load for generation or clock time for generation respectively, that scales linearly with the number of weak learners.
For decoding, once the token-level distributions generated by the weak learners are combined using EmsemW2S algorithm, efficient decoding algorithms like speculative decoding can be employed to produce the final response. However, this is not the focus of this work.

We do agree that the primary increase in computational cost compared to the baseline---where a single weak expert supervises the student---comes from multi-LLM decoding. However, its important to consider that unlike other multi-LLM works \cite{anonymous2025collab, du2023improving}, we do not use an additional reward or judge model for combining multi-LLM outputs, significantly reducing hardware demands. Moreover, we rely on the same labeled data rather than collecting new data to train separate models. For this same reason we use datasets which do not have very long outputs and are option based.

\section{Results}\label{app:exp}

\subsection{Baselines for weak-expert performance comparison}\label{appx:weak_model}

In this section we show one additional baseline to compare weak model's performance. Basically instead of using adaptive weights of the model we use fixed model weights i.e. $\alpha_t = 1 \forall t = [1,T]$. This is similar to bagging. You can find the results in the table \ref{tab:weak_baseline}.

\begin{table}[htp]
\resizebox{\columnwidth}{!}{%
\begin{tabular}{l|llllllll}
\hline
                                                                & \multicolumn{4}{l}{Qwen-1.5B}                     & \multicolumn{4}{l}{Qwen-3B}                       \\ \cline{2-9} 
 & \multicolumn{2}{l}{Random (ID)} & \multicolumn{2}{l}{Easy-Hard (OOD)} & \multicolumn{2}{l}{Random (ID)} & \multicolumn{2}{l}{Easy-Hard (OOD)} \\ \cline{2-9} 
                                                                & Single Model & 5 Models & Single Model & 5 Models & Single Model & 5 Models & Single Model & 5 Models \\ \hline
ARC                                                             & 46.02        &   47.00       & 38.14        & 38.22    & 54.02        &       52.75   & 42.92        & 43.00    \\
Quartz                                                          & 74.23        &    74.36      & 72.83        & 71.55    & 81.25        &    76.02      & 80.99        & 76.02    \\
Math                                                            & 47.17        &     47.20     & 45.42        &   46.00       & 61.74        &     56.3     & 50.16        &   50.02       \\
\hline
\end{tabular}%
}
\caption{In this table we compare performance of weak experts on additional baseline where we combine all the model (specifically 5 models) by giving each of them equal weight. The column 5 models depicts that. We also mention the baseline shown in the main table which uses the first model. We observe that for all the case the 5-model baseline perform worse or similar to the single model baseline.}
\label{tab:weak_baseline}
\end{table}

\subsection{Results on ARC Dataset for generation task}\label{appx:arc_results}

\begin{figure}[H]
  \begin{minipage}{\textwidth}
  \includegraphics[trim={0 0 0 0},clip,width=1\linewidth]{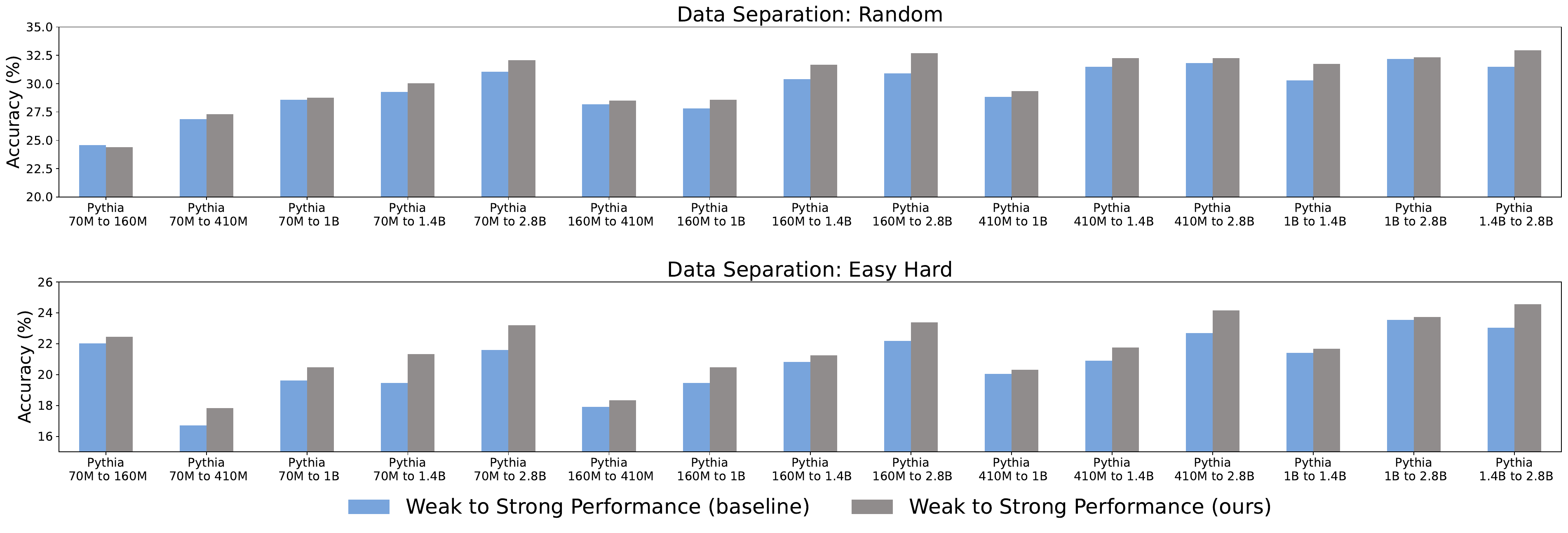}
  \end{minipage}
  \begin{minipage}{\textwidth}
    \includegraphics[trim={0 8cm 0 8cm},clip,width=\linewidth]{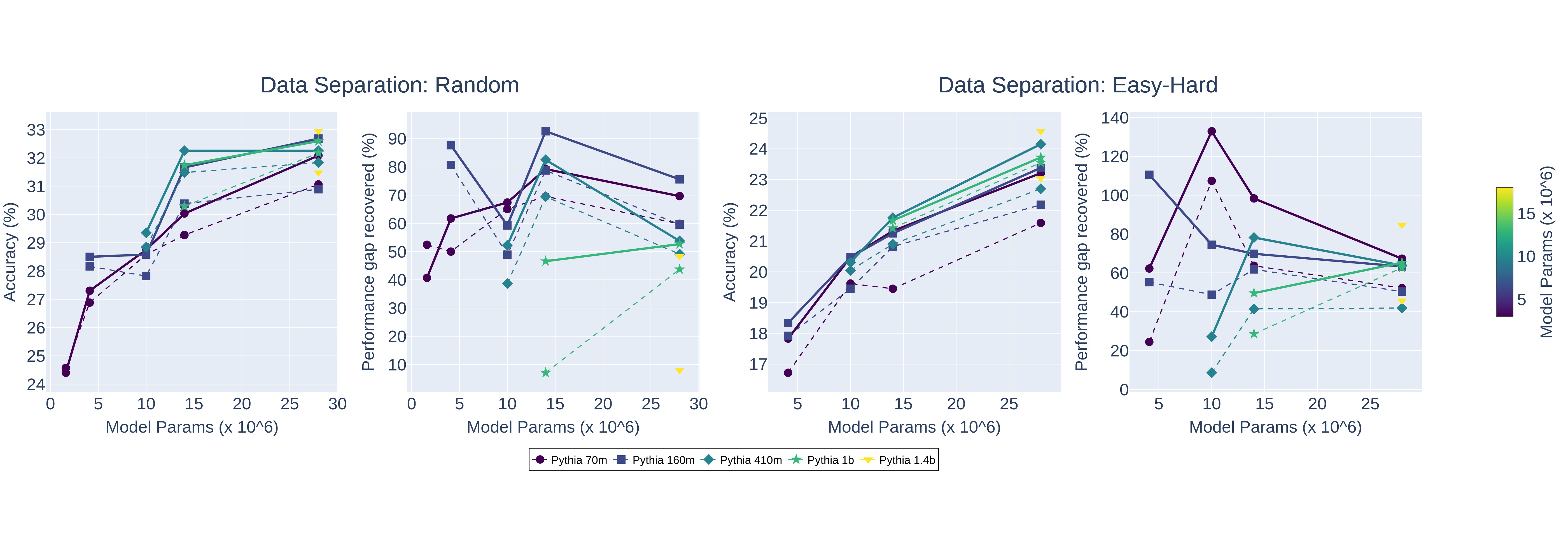}
  \end{minipage}
  \vspace{-1em}
\caption{\textbf{Generation Task (ARC Data):} \textbf{Top figure} shows a bar plot comparing the w2s generalization of our method (grey) with a baseline (blue) for various combinations of weak and strong model pairs for the SFT task on Q/A data for random data split (top bar-plot) and easy-hard split (bottom bar-plot). \textbf{Bottom figure} shows a line plot comparing accuracy and PGR. The left two figures are for random data split, while the right two are for the easy-hard split to show e2h generalization.}
\label{fig:bar_2_arc}
\end{figure}

\subsection{Results on Sciq Dataset for Binary Classification Task}\label{appx:binary_results}

\begin{figure*}[htp]
    \centering
    \begin{minipage}[t]{0.9\textwidth}
        \includegraphics[trim={0 0 0 0},clip,width=1\linewidth]{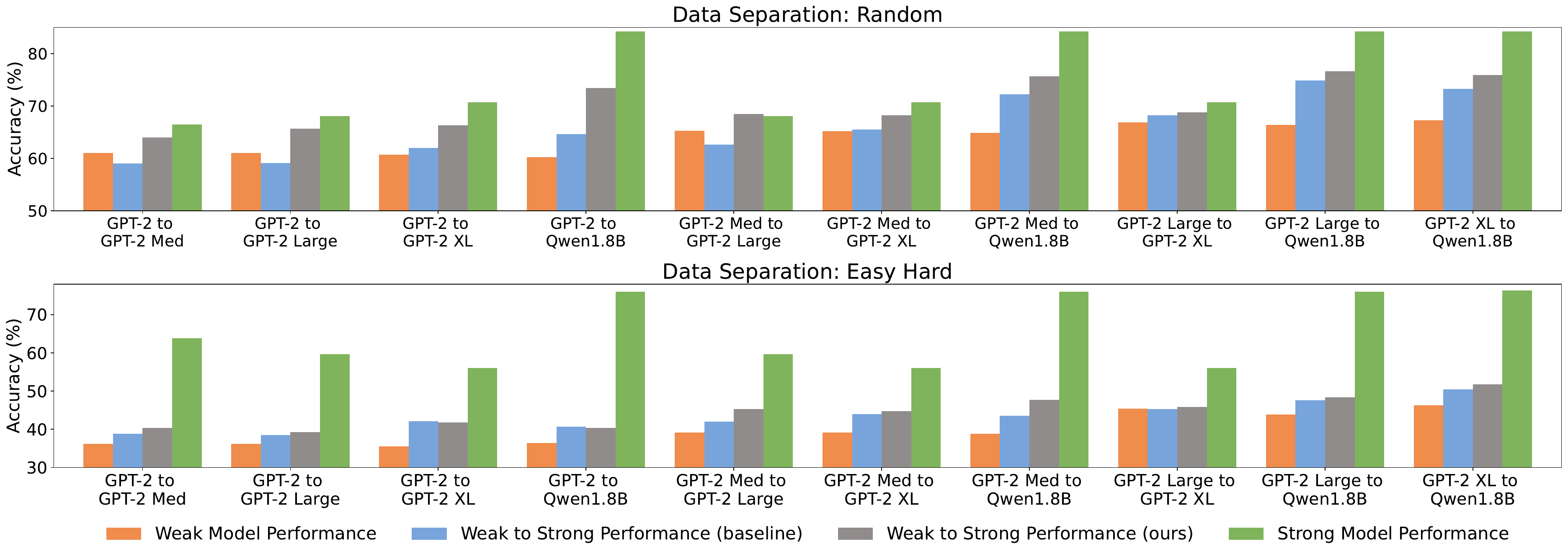}
    \end{minipage}
    \hfill
    \begin{minipage}[t]{0.9\textwidth}
        \includegraphics[trim={0 8cm 0 8cm},clip,width=\linewidth]{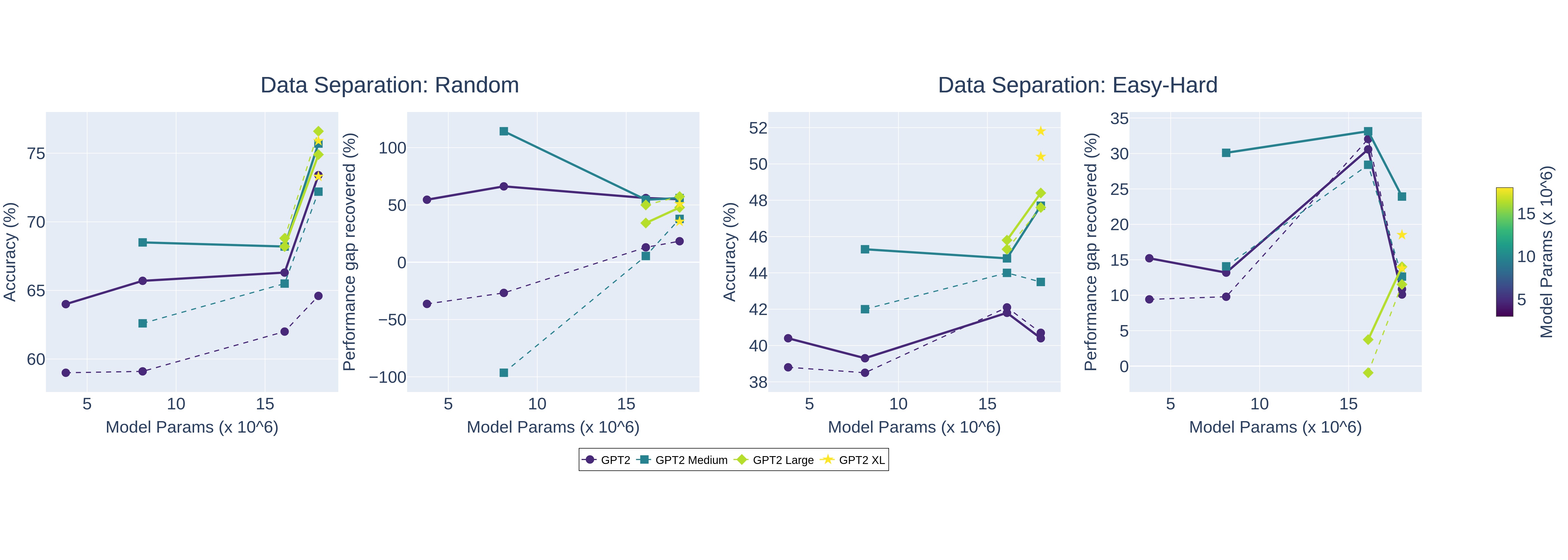}
    \end{minipage}
    \caption[\textwidth]{\textbf{Binary Classification Task.} \textbf{Top figure} shows a bar plot comparing w2s generalization of our method (grey)  with a baseline (blue) from \cite{openAIw2s} using accuracy values(\%) for different combinations of weak and strong model pairs for random data split (top bar-plot) and easy-hard split(bottom bar-plot). \textbf{Bottom figure} shows a line plot comparing the accuracy and performance gap recovered values (PGR). The left two figures are for random data split, while the right two figures are for the easy-hard split to show e2h generalization.}
    \label{fig:binary_plots}
    \vspace{-10pt}
\end{figure*}

\textbf{W2S Results with Random Training Data Splits.} The baseline of this method is a replication of \cite{openAIw2s}. From Figure \ref{fig:binary_plots}, by applying AdaBoost, we observe a significant improvement in the weak model accuracy, significantly improving the PGR values. In the case of the GPT-2-medium to GPT-2-large pair, we even see the PGR exceeding 100\%, meaning that the transfer model has outperformed the strong model's performance. This is the ambitious aim of the w2s generalization problem, and our results show that w2s generalization is achievable.

\textbf{W2S Results with Easy and Hard Training Data Splits.}
From Figure \ref{fig:binary_plots}, we see that applying AdaBoost significantly improves weak model accuracy, thereby enhancing the PGR values. However, for this holistic e2h generalization problem, we are far from reaching the full capability of a strong model. For very small (GPT-2) and large model pairs (GPT-2-xl and above), we do not see improvement in w2s generalization despite the weak models' accuracy improvements. Overall, we observe an improvement of up to 14\% in accuracy compared to the baseline and an average improvement of 6.52\% and 3\% for random and easy-hard splits, respectively.

\textbf{Scaling Law.} In Figure \ref{fig:binary_plots} (line plot), we see less PGR recovery for the Qwen-1.8B model even though it is similar in size to GPT-2-xl. Similarly, in the bar plot, we see a drastic difference between the oracle performance of GPT2xl and Qwen-1.8B. This is because the Qwen models series are more capable even after being the same size. Thus, model size is not a good metric, but model capability is a better metric for differentiating between weak and strong models.

\section{Aggregated plots}
\begin{figure}[H]
 \centering
  \includegraphics[trim={0 0 0 0},clip,width=1\linewidth]{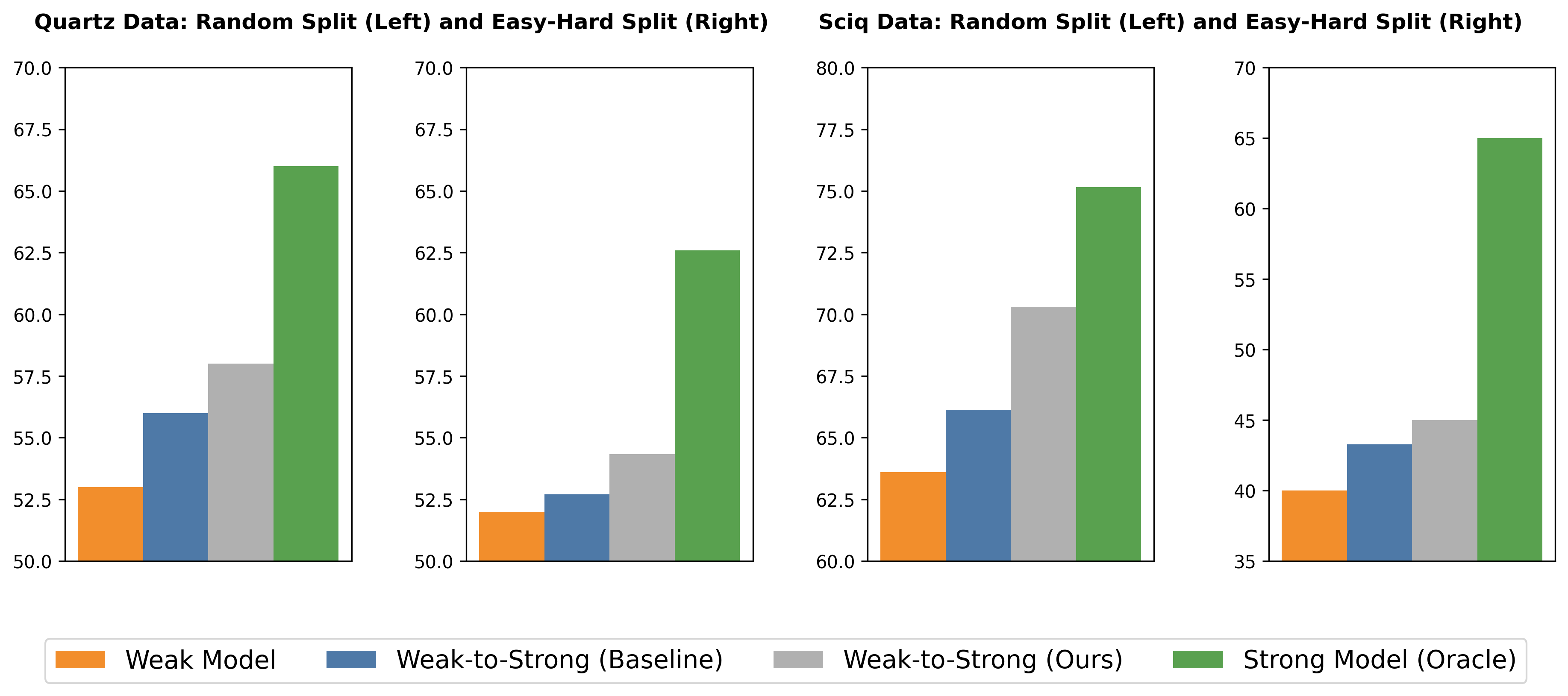}
\caption{\textbf{Aggregated results for Quartz Data on Generation Task and Sciq Data on Binary Classification Task} for both random and easy-hard data splits. We aggregate results for three experimental runs will different seed across all model pairs similar to \cite{openAIw2s}.}
\label{fig:aggregated_plots}
\end{figure}

\newpage
\section{Generative Task Details}

\subsection{ Supervised-Fine Tuning task for Quartz Question-Answer Dataset}

\begin{table}[H]
\resizebox{\textwidth}{!}{%
\begin{tabular}{llllllll}
                     & \multicolumn{3}{l}{\cellcolor[HTML]{9698ED}Weak Model}  & \cellcolor[HTML]{9698ED} & \multicolumn{3}{l}{\cellcolor[HTML]{9698ED}Strong Model} \\
                     & Token-Avg Acc    & Option Acc     & Option Acc(on w2s)  & $\alpha$                 & oracle         & Token-Avg Acc  & Option Acc             \\
                     & \multicolumn{3}{l}{\cellcolor[HTML]{C0C0C0}Pythia-70m}  & \cellcolor[HTML]{EFEFEF} & \multicolumn{3}{l}{\cellcolor[HTML]{EFEFEF}Pythia-160m}  \\
Baseline             & 17.95 ± 0.44     & 50.21 ± 0.23   & 49.7 ± 0.28         & 10.81 ± 0.04             & 50.77 ± 0.26   & 34.3 ± 0.44    & 51.11 ± 0.23           \\
With Adaboost (T:03) & 25.94 ± 0.38     & 50.64 ± 0.39   & 49.43 ± 0.25        & 10.67 ± 0.05             & 50.77 ± 0.26   & 34.17 ± 0.36   & \textbf{51.66 ± 0.45}  \\
                     & \multicolumn{3}{l}{\cellcolor[HTML]{C0C0C0}Pythia-70m}  & \cellcolor[HTML]{EFEFEF} & \multicolumn{3}{l}{\cellcolor[HTML]{EFEFEF}Pythia-410m}  \\
Baseline             & 17.95 ± 0.44     & 50.21 ± 0.23   & 49.7 ± 0.28         & 10.81 ± 0.04             & 59.18 ± 0.78   & 50.28 ± 0.44   & 50.68 ± 0.3            \\
With Adaboost (T:04) & 25.22 ± 0.15     & 50.51 ± 0.53   & 49.8 ± 0.14         & 10.68 ± 0.05             & 59.18 ± 0.78   & 50.88 ± 0.18   & \textbf{52.42 ± 0.33}  \\
                     & \multicolumn{3}{l}{\cellcolor[HTML]{C0C0C0}Pythia-70m}  & \cellcolor[HTML]{EFEFEF} & \multicolumn{3}{l}{\cellcolor[HTML]{EFEFEF}Pythia-1b}    \\
Baseline             & 17.95 ± 0.44     & 50.21 ± 0.23   & 49.7 ± 0.28         & 10.81 ± 0.04             & 63.35 ± 0.3    & 51.87 ± 0.11   & 50.89 ± 0.16           \\
With Adaboost (T:05) & 26.2 ± 0.06      & 50.55 ± 0.28   & 49.65 ± 0.11        & 10.66 ± 0.04             & 63.35 ± 0.3    & 51.83 ± 0.38   & \textbf{51.83 ± 0.31}  \\
                     & \multicolumn{3}{l}{\cellcolor[HTML]{C0C0C0}Pythia-70m}  & \cellcolor[HTML]{EFEFEF} & \multicolumn{3}{l}{\cellcolor[HTML]{EFEFEF}Pythia-1.4b}  \\
Baseline             & 17.89 ± 0.46     & 49.87 ± 0.06   & 49.46 ± 0.35        & 10.82 ± 0.05             & 68.83 ± 1.28   & 51.82 ± 0.05   & 50.17 ± 0.24           \\
With Adaboost (T:04) & 25.32 ± 0.82     & 50.04 ± 0.37   & 49.23 ± 0.27        & 10.7 ± 0.06              & 68.83 ± 1.28   & 51.76 ± 0.17   & \textbf{51.45 ± 0.07}  \\
                     & \multicolumn{3}{l}{\cellcolor[HTML]{C0C0C0}Pythia-70m}  & \cellcolor[HTML]{EFEFEF} & \multicolumn{3}{l}{\cellcolor[HTML]{EFEFEF}Pythia-2.8b}  \\
Baseline             & 18.06 ± 0.39     & 49.4 ± 0.39    & 49.73 ± 0.33        & 10.86 ± 0.02             & 73.38 ± 1.02   & 52.28 ± 0.29   & 50.21 ± 0.23           \\
With Adaboost (T:02) & 24.37 ± 0.99     & 50.13 ± 0.4    & 49.48 ± 0.21        & 10.74 ± 0.04             & 73.38 ± 1.02   & 52.3 ± 0.14    & \textbf{51.02 ± 0.22}  \\
                     & \multicolumn{3}{l}{\cellcolor[HTML]{C0C0C0}Pythia-160m} & \cellcolor[HTML]{EFEFEF} & \multicolumn{3}{l}{\cellcolor[HTML]{EFEFEF}Pythia-410m}  \\
Baseline             & 33.51 ± 0.19     & 50.81 ± 1.0    & 49.6 ± 0.27         & 10.03 ± 0.0              & 59.18 ± 0.78   & 50.39 ± 0.3    & 50.68 ± 0.5            \\
With Adaboost (T:04) & 40.85 ± 0.49     & 51.79 ± 0.48   & 49.08 ± 0.32        & 9.81 ± 0.05              & 59.18 ± 0.78   & 50.39 ± 0.18   & \textbf{52.13 ± 0.3}   \\
                     & \multicolumn{3}{l}{\cellcolor[HTML]{C0C0C0}Pythia-160m} & \cellcolor[HTML]{EFEFEF} & \multicolumn{3}{l}{\cellcolor[HTML]{EFEFEF}Pythia-1b}    \\
Baseline             & 33.51 ± 0.19     & 50.81 ± 1.0    & 49.6 ± 0.27         & 10.03 ± 0.0              & 63.35 ± 0.3    & 52.36 ± 0.29   & 50.6 ± 0.33            \\
With Adaboost (T:02) & 40.61 ± 0.8      & 51.36 ± 0.25   & 49.93 ± 0.52        & 9.76 ± 0.05              & 63.35 ± 0.3    & 52.45 ± 0.42   & \textbf{51.92 ± 0.31}  \\
                     & \multicolumn{3}{l}{\cellcolor[HTML]{C0C0C0}Pythia-160m} & \cellcolor[HTML]{EFEFEF} & \multicolumn{3}{l}{\cellcolor[HTML]{EFEFEF}Pythia-1.4b}  \\
Baseline             & 33.42 ± 0.23     & 51.4 ± 0.59    & 49.43 ± 0.41        & 10.03 ± 0.0              & 68.83 ± 1.28   & 52.02 ± 0.2    & 51.02 ± 0.55           \\
With Adaboost (T:03) & 40.87 ± 0.49     & 51.02 ± 0.18   & 49.28 ± 0.13        & 9.75 ± 0.02              & 68.83 ± 1.28   & 52.11 ± 0.39   & \textbf{53.02 ± 0.55}  \\
                     & \multicolumn{3}{l}{\cellcolor[HTML]{C0C0C0}Pythia-160m} & \cellcolor[HTML]{EFEFEF} & \multicolumn{3}{l}{\cellcolor[HTML]{EFEFEF}Pythia-2.8b}  \\
Baseline             & 33.42 ± 0.23     & 51.4 ± 0.59    & 49.43 ± 0.41        & 10.03 ± 0.0              & 73.17 ± 0.88   & 52.82 ± 0.02   & 51.45 ± 0.5            \\
With Adaboost (T:04) & 41.13 ± 0.51     & 51.23 ± 0.4    & 49.65 ± 0.14        & 9.78 ± 0.06              & 73.17 ± 0.88   & 52.51 ± 0.3    & \textbf{51.74 ± 0.17}  \\
                     & \multicolumn{3}{l}{\cellcolor[HTML]{C0C0C0}Pythia-410m} & \cellcolor[HTML]{EFEFEF} & \multicolumn{3}{l}{\cellcolor[HTML]{EFEFEF}Pythia-1b}    \\
Baseline             & 52.71 ± 0.24     & 59.27 ± 0.46   & 55.54 ± 0.49        & 10.0 ± 0.01              & 63.35 ± 0.3    & 53.39 ± 0.2    & 56.21 ± 0.76           \\
With Adaboost (T:02) & 53.39 ± 0.17     & 58.5 ± 0.33    & 55.91 ± 0.35        & 9.69 ± 0.08              & 63.35 ± 0.3    & 53.87 ± 0.46   & \textbf{56.42 ± 0.56}  \\
                     & \multicolumn{3}{l}{\cellcolor[HTML]{C0C0C0}Pythia-410m} & \cellcolor[HTML]{EFEFEF} & \multicolumn{3}{l}{\cellcolor[HTML]{EFEFEF}Pythia-1.4b}  \\
Baseline             & 52.9 ± 0.09      & 59.65 ± 0.15   & 55.66 ± 0.51        & 9.98 ± 0.02              & 68.83 ± 1.28   & 53.33 ± 0.74   & 56.34 ± 0.9            \\
With Adaboost (T:02) & 53.26 ± 0.27     & 58.8 ± 0.42    & 56.11 ± 0.34        & 9.66 ± 0.08              & 68.83 ± 1.28   & 54.14 ± 0.63   & \textbf{57.7 ± 0.61}   \\
                     & \multicolumn{3}{l}{\cellcolor[HTML]{C0C0C0}Pythia-410m} & \cellcolor[HTML]{EFEFEF} & \multicolumn{3}{l}{\cellcolor[HTML]{EFEFEF}Pythia-2.8b}  \\
Baseline             & 52.13 ± 0.64     & 58.29 ± 1.1    & 55.94 ± 0.3         & 9.89 ± 0.06              & 73.38 ± 1.02   & 54.38 ± 0.31   & 55.74 ± 0.73           \\
With Adaboost (T:04) & 53.39 ± 0.19     & 59.18 ± 0.42   & 55.32 ± 0.51        & 9.85 ± 0.05              & 73.38 ± 1.02   & 55.71 ± 0.53   & \textbf{59.01 ± 0.94}  \\
                     & \multicolumn{3}{l}{\cellcolor[HTML]{C0C0C0}Pythia-1b}   & \cellcolor[HTML]{EFEFEF} & \multicolumn{3}{l}{\cellcolor[HTML]{EFEFEF}Pythia-1.4b}  \\
Baseline             & 55.65 ± 0.52     & 61.99 ± 0.51   & 58.6 ± 1.13         & 9.85 ± 0.01              & 68.62 ± 0.12   & 55.33 ± 0.31   & 58.93 ± 0.68           \\
With Adaboost (T:03) & 56.81 ± 0.47     & 62.12 ± 0.43   & 58.14 ± 0.85        & 9.74 ± 0.11              & 68.62 ± 0.12   & 55.99 ± 0.16   & \textbf{61.69 ± 0.57}  \\
                     & \multicolumn{3}{l}{\cellcolor[HTML]{C0C0C0}Pythia-1b}   & \cellcolor[HTML]{EFEFEF} & \multicolumn{3}{l}{\cellcolor[HTML]{EFEFEF}Pythia-2.8b}  \\
Baseline             & 55.54 ± 0.6      & 62.12 ± 0.51   & 58.55 ± 1.14        & 9.84 ± 0.01              & 73.3 ± 0.3     & 57.26 ± 0.3    & 61.52 ± 1.38           \\
With Adaboost (T:02) & 57.09 ± 0.41     & 62.84 ± 0.12   & 59.0 ± 0.62         & 9.63 ± 0.02              & 73.3 ± 0.3     & 58.1 ± 0.08    & \textbf{63.99 ± 0.93}  \\
                     & \multicolumn{3}{l}{\cellcolor[HTML]{C0C0C0}Pythia-1.4b} & \cellcolor[HTML]{EFEFEF} & \multicolumn{3}{l}{\cellcolor[HTML]{EFEFEF}Pythia-2.8b}  \\
Baseline             & 57.11 ± 0.45     & 69.64 ± 0.97   & 66.87 ± 1.1         & 9.87 ± 0.02              & 73.76 ± 0.67   & 59.34 ± 0.24   & 67.94 ± 0.78           \\
With Adaboost (T:02) & 59.17 ± 0.12     & 70.66 ± 0.06   & 67.29 ± 0.77        & 9.65 ± 0.03              & 73.76 ± 0.67   & 59.3 ± 0.34    & \textbf{68.92 ± 1.06} 
\end{tabular}%
}
\caption{This table shows weak to strong generalization using random data-splits for quartz dataset. We also study the impact of using ensemble learning methods, which combines weak learners, for weak to strong training. Each model is trained for 5 epochs and uses a learning rate of $5x10^{-5}$. The values in this table are generated by aggregating 3 experiments. We show here mean and Standard Error of the Mean values.}
\label{tab:quartz_random}
\end{table}

\begin{figure}[htp]
 \centering
  \includegraphics[trim={0 0 0 0},clip,width=1\linewidth]{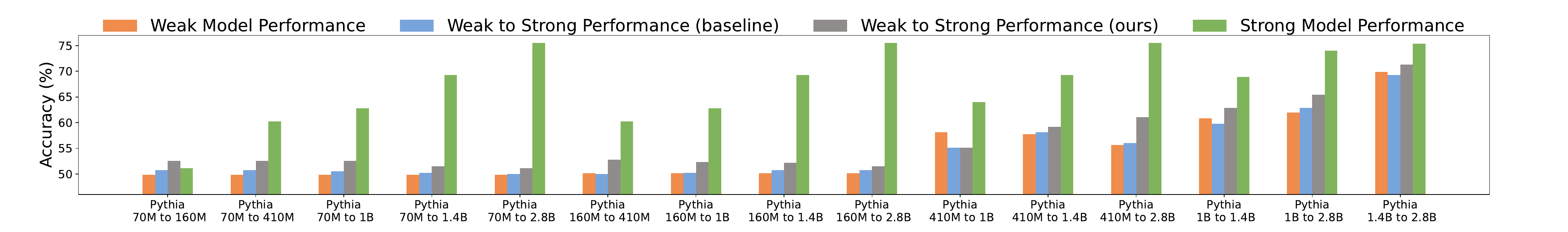}
\caption{\textbf{Quartz Dataset (Random):} This figure shows bar plots comparing accuracy values of weak model performance, w2s model performance (baseline and ours) and strong model performance (oracle) for one specific run of experiments. Values are also mentioned in table \ref{tab:quartz_improv}.}
\label{fig:bar_random_quartz}
\end{figure}

\begin{table}[H]
\resizebox{\textwidth}{!}{%
\begin{tabular}{llllllll}
                     & \multicolumn{3}{l}{\cellcolor[HTML]{9698ED}Weak Model}  & \cellcolor[HTML]{9698ED} & \multicolumn{3}{l}{\cellcolor[HTML]{9698ED}Strong Model} \\
                     & Token-Avg Acc    & Option Acc     & Option Acc(on w2s)  & $\alpha$                 & oracle         & Token-Avg Acc  & Option Acc             \\
                     & \multicolumn{3}{l}{\cellcolor[HTML]{C0C0C0}Pythia-70m}  & \cellcolor[HTML]{EFEFEF} & \multicolumn{3}{l}{\cellcolor[HTML]{EFEFEF}Pythia-160m}  \\
Baseline             & 16.27 ± 0.14     & 48.0 ± 0.51    & 49.21 ± 0.05        & 10.53 ± 0.0              & 47.11 ± 0.28   & 29.24 ± 0.18   & 49.11 ± 0.39           \\
With Adaboost (T:03) & 23.31 ± 0.9      & 47.11 ± 0.31   & 49.23 ± 0.41        & 10.43 ± 0.03             & 47.11 ± 0.28   & 29.24 ± 0.25   & \textbf{49.32 ± 0.23}  \\
                     & \multicolumn{3}{l}{\cellcolor[HTML]{C0C0C0}Pythia-70m}  & \cellcolor[HTML]{EFEFEF} & \multicolumn{3}{l}{\cellcolor[HTML]{EFEFEF}Pythia-410m}  \\
Baseline             & 16.27 ± 0.14     & 48.0 ± 0.51    & 49.21 ± 0.05        & 10.53 ± 0.0              & 52.3 ± 0.39    & 43.63 ± 0.29   & 47.32 ± 0.36           \\
With Adaboost (T:04) & 23.81 ± 1.01     & 47.66 ± 0.5    & 49.06 ± 0.2         & 10.42 ± 0.02             & 52.3 ± 0.39    & 43.53 ± 0.44   & \textbf{48.13 ± 0.47}  \\
                     & \multicolumn{3}{l}{\cellcolor[HTML]{C0C0C0}Pythia-70m}  & \cellcolor[HTML]{EFEFEF} & \multicolumn{3}{l}{\cellcolor[HTML]{EFEFEF}Pythia-1b}    \\
Baseline             & 16.27 ± 0.14     & 48.0 ± 0.51    & 49.21 ± 0.05        & 10.53 ± 0.0              & 55.91 ± 0.37   & 47.48 ± 0.23   & 47.92 ± 0.23           \\
With Adaboost (T:05) & 24.64 ± 0.22     & 47.49 ± 0.49   & 49.41 ± 0.38        & 10.39 ± 0.0              & 55.91 ± 0.37   & 45.5 ± 0.74    & \textbf{49.74 ± 0.24}  \\
                     & \multicolumn{3}{l}{\cellcolor[HTML]{C0C0C0}Pythia-70m}  & \cellcolor[HTML]{EFEFEF} & \multicolumn{3}{l}{\cellcolor[HTML]{EFEFEF}Pythia-1.4b}  \\
Baseline             & 16.07 ± 0.22     & 48.17 ± 0.43   & 49.38 ± 0.14        & 10.58 ± 0.04             & 65.35 ± 0.66   & 46.25 ± 0.61   & 47.96 ± 0.34           \\
With Adaboost (T:04) & 23.79 ± 0.55     & 46.94 ± 0.18   & 49.58 ± 0.27        & 10.44 ± 0.04             & 65.35 ± 0.66   & 45.53 ± 0.2    & \textbf{50.68 ± 0.17}  \\
                     & \multicolumn{3}{l}{\cellcolor[HTML]{C0C0C0}Pythia-70m}  & \cellcolor[HTML]{EFEFEF} & \multicolumn{3}{l}{\cellcolor[HTML]{EFEFEF}Pythia-2.8b}  \\
Baseline             & 16.12 ± 0.21     & 48.85 ± 0.48   & 49.75 ± 0.32        & 10.63 ± 0.04             & 70.2 ± 0.17    & 48.08 ± 0.18   & 48.85 ± 0.31           \\
With Adaboost (T:02) & 22.96 ± 0.75     & 47.02 ± 0.12   & 49.36 ± 0.11        & 10.5 ± 0.05              & 70.2 ± 0.17    & 48.58 ± 0.16   & \textbf{49.87 ± 0.06}  \\
                     & \multicolumn{3}{l}{\cellcolor[HTML]{C0C0C0}Pythia-160m} & \cellcolor[HTML]{EFEFEF} & \multicolumn{3}{l}{\cellcolor[HTML]{EFEFEF}Pythia-410m}  \\
Baseline             & 25.61 ± 0.33     & 47.75 ± 0.35   & 49.83 ± 0.29        & 9.96 ± 0.02              & 52.3 ± 0.39    & 42.75 ± 0.91   & 47.75 ± 0.61           \\
With Adaboost (T:04) & 29.63 ± 0.55     & 47.02 ± 0.09   & 48.47 ± 0.3         & 9.7 ± 0.09               & 52.3 ± 0.39    & 43.78 ± 0.14   & \textbf{48.42 ± 0.12}  \\
                     & \multicolumn{3}{l}{\cellcolor[HTML]{C0C0C0}Pythia-160m} & \cellcolor[HTML]{EFEFEF} & \multicolumn{3}{l}{\cellcolor[HTML]{EFEFEF}Pythia-1b}    \\
Baseline             & 25.61 ± 0.33     & 47.75 ± 0.35   & 49.83 ± 0.29        & 9.96 ± 0.02              & 55.91 ± 0.37   & 46.08 ± 0.38   & 49.36 ± 0.53           \\
With Adaboost (T:02) & 28.96 ± 0.23     & 46.43 ± 0.18   & 48.49 ± 0.11        & 9.69 ± 0.09              & 55.91 ± 0.37   & 44.7 ± 0.58    & \textbf{49.15 ± 0.73}  \\
                     & \multicolumn{3}{l}{\cellcolor[HTML]{C0C0C0}Pythia-160m} & \cellcolor[HTML]{EFEFEF} & \multicolumn{3}{l}{\cellcolor[HTML]{EFEFEF}Pythia-1.4b}  \\
Baseline             & 25.76 ± 0.43     & 47.15 ± 0.15   & 49.26 ± 0.2         & 9.96 ± 0.02              & 65.35 ± 0.66   & 45.83 ± 0.64   & 49.7 ± 0.85            \\
With Adaboost (T:03) & 28.83 ± 0.84     & 46.56 ± 0.27   & 48.17 ± 0.14        & 9.64 ± 0.06              & 65.35 ± 0.66   & 45.4 ± 0.44    & \textbf{50.0 ± 0.22}   \\
                     & \multicolumn{3}{l}{\cellcolor[HTML]{C0C0C0}Pythia-160m} & \cellcolor[HTML]{EFEFEF} & \multicolumn{3}{l}{\cellcolor[HTML]{EFEFEF}Pythia-2.8b}  \\
Baseline             & 26.46 ± 0.25     & 47.49 ± 0.33   & 48.98 ± 0.14        & 10.02 ± 0.03             & 70.2 ± 0.17    & 48.03 ± 0.13   & 49.4 ± 0.3             \\
With Adaboost (T:04) & 29.61 ± 0.51     & 46.6 ± 0.25    & 48.69 ± 0.47        & 9.54 ± 0.03              & 70.2 ± 0.17    & 48.4 ± 0.29    & \textbf{50.3 ± 0.41}   \\
                     & \multicolumn{3}{l}{\cellcolor[HTML]{C0C0C0}Pythia-410m} & \cellcolor[HTML]{EFEFEF} & \multicolumn{3}{l}{\cellcolor[HTML]{EFEFEF}Pythia-1b}    \\
Baseline             & 36.73 ± 0.39     & 51.06 ± 0.39   & 53.26 ± 0.38        & 10.07 ± 0.01             & 55.91 ± 0.37   & 46.6 ± 0.38    & 50.72 ± 0.68           \\
With Adaboost (T:02) & 38.11 ± 0.44     & 49.36 ± 0.21   & 51.66 ± 0.35        & 9.76 ± 0.14              & 55.91 ± 0.37   & 46.4 ± 0.35    & \textbf{52.09 ± 0.3}   \\
                     & \multicolumn{3}{l}{\cellcolor[HTML]{C0C0C0}Pythia-410m} & \cellcolor[HTML]{EFEFEF} & \multicolumn{3}{l}{\cellcolor[HTML]{EFEFEF}Pythia-1.4b}  \\
Baseline             & 37.23 ± 0.27     & 51.11 ± 0.4    & 53.19 ± 0.42        & 10.04 ± 0.03             & 65.35 ± 0.66   & 47.73 ± 0.78   & 53.66 ± 0.56           \\
With Adaboost (T:02) & 38.31 ± 0.23     & 50.17 ± 0.44   & 51.56 ± 0.22        & 9.53 ± 0.09              & 65.35 ± 0.66   & 48.35 ± 0.18   & \textbf{53.36 ± 0.5}   \\
                     & \multicolumn{3}{l}{\cellcolor[HTML]{C0C0C0}Pythia-410m} & \cellcolor[HTML]{EFEFEF} & \multicolumn{3}{l}{\cellcolor[HTML]{EFEFEF}Pythia-2.8b}  \\
Baseline             & 37.13 ± 0.23     & 51.02 ± 0.47   & 52.87 ± 0.21        & 10.03 ± 0.03             & 70.2 ± 0.17    & 48.48 ± 0.36   & 54.47 ± 0.16           \\
With Adaboost (T:04) & 38.13 ± 0.26     & 49.87 ± 0.68   & 51.49 ± 0.28        & 9.6 ± 0.04               & 70.2 ± 0.17    & 49.05 ± 0.14   & \textbf{55.36 ± 0.47}  \\
                     & \multicolumn{3}{l}{\cellcolor[HTML]{C0C0C0}Pythia-1b}   & \cellcolor[HTML]{EFEFEF} & \multicolumn{3}{l}{\cellcolor[HTML]{EFEFEF}Pythia-1.4b}  \\
Baseline             & 40.3 ± 0.46      & 54.51 ± 0.73   & 54.25 ± 0.26        & 10.33 ± 0.08             & 66.67 ± 0.72   & 47.0 ± 0.22    & 56.76 ± 0.58           \\
With Adaboost (T:03) & 40.75 ± 0.67     & 53.36 ± 0.92   & 53.61 ± 0.44        & 11.0 ± 0.72              & 66.67 ± 0.72   & 47.25 ± 0.32   & \textbf{57.23 ± 0.37}  \\
                     & \multicolumn{3}{l}{\cellcolor[HTML]{C0C0C0}Pythia-1b}   & \cellcolor[HTML]{EFEFEF} & \multicolumn{3}{l}{\cellcolor[HTML]{EFEFEF}Pythia-2.8b}  \\
Baseline             & 40.33 ± 0.44     & 54.08 ± 1.07   & 54.33 ± 0.19        & 10.33 ± 0.08             & 73.09 ± 0.42   & 49.2 ± 0.2     & 58.08 ± 0.38           \\
With Adaboost (T:02) & 40.53 ± 0.34     & 52.34 ± 0.09   & 53.39 ± 0.2         & 11.68 ± 0.75             & 73.09 ± 0.42   & 49.48 ± 0.3    & \textbf{59.35 ± 0.52}  \\
                     & \multicolumn{3}{l}{\cellcolor[HTML]{C0C0C0}Pythia-1.4b} & \cellcolor[HTML]{EFEFEF} & \multicolumn{3}{l}{\cellcolor[HTML]{EFEFEF}Pythia-2.8b}  \\
Baseline             & 42.2 ± 1.12      & 59.69 ± 0.83   & 62.39 ± 1.06        & 10.3 ± 0.1               & 73.17 ± 0.38   & 51.22 ± 0.5    & 62.46 ± 0.91           \\
With Adaboost (T:02) & 42.98 ± 0.64     & 59.82 ± 0.51   & 61.38 ± 0.48        & 10.52 ± 0.35             & 73.17 ± 0.38   & 51.72 ± 0.37   & \textbf{63.01 ± 0.28} 
\end{tabular}%
}
\caption{This table shows weak to strong generalization using easy-hard data-splits for quartz dataset. We also study the impact of using ensemble learning methods, which combines weak learners, for weak to strong training. Each model is trained for 5 epochs and uses a learning rate of $5\times10^{-5}$. The values in this table are generated by aggregating 3 experiments. We show here mean and Standard Error of the Mean values.}
\label{tab:quartz_easy}
\end{table}

\begin{figure}[htp]
 \centering
  \includegraphics[trim={0 0 0 0},clip,width=1\linewidth]{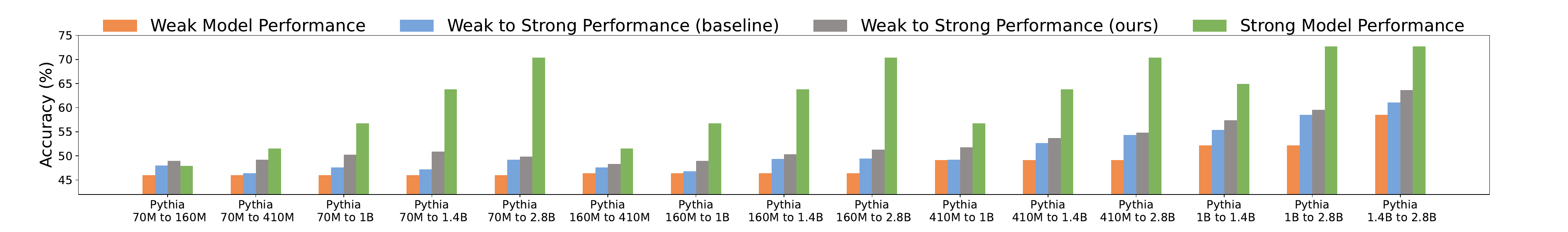}
\caption{\textbf{Quartz Dataset (Easy-Hard):} This figure shows bar plots comparing accuracy values of weak model performance, w2s model performance (baseline and ours) and strong model performance (oracle) for one specific run of experiments. Values are also mentioned in table \ref{tab:quartz_improv}.}
\label{fig:bar_easy_quartz}
\end{figure}

\begin{table}[H]
\resizebox{\textwidth}{!}{%
\begin{tabular}{|l|l|llll|l|llll|r|}
\hline
\multirow{3}{*}{\begin{tabular}[c]{@{}l@{}}Weak \\ Model \\ Size\end{tabular}} & \multirow{3}{*}{\begin{tabular}[c]{@{}l@{}}Strong \\ Model \\ Size\end{tabular}} & \multicolumn{2}{l|}{Data Separation: Random} & \multirow{3}{*}{Improv(\%)} & \multicolumn{2}{l|}{Data Separation: Easy-Hard} & \multicolumn{1}{l|}{\multirow{3}{*}{Improv(\%)}} \\ \cline{3-4} \cline{6-7}
 &  & \multicolumn{2}{l|}{W2S Performance} &  & \multicolumn{2}{l|}{W2S Performance} & \multicolumn{1}{l|}{} \\ \cline{3-4} \cline{6-7}
 &  & \multicolumn{1}{l|}{Baseline} & Ours &  & \multicolumn{1}{l|}{Baseline} & Ours & \multicolumn{1}{l|}{} \\ \hline
Pythia-70M & Pythia-160M & \multicolumn{1}{l|}{0.5077} & 0.5255 & 3.5\% & \multicolumn{1}{l|}{0.48} & 0.4898 & 2\% \\ \hline
Pythia-70M & Pythia-410M & \multicolumn{1}{l|}{0.5077} & 0.5255 & 3.5\% & \multicolumn{1}{l|}{0.4643} & 0.4923 & 6\% \\ \hline
Pythia-70M & Pythia-1B & \multicolumn{1}{l|}{0.5051} & 0.5255 & 4\% & \multicolumn{1}{l|}{0.4758} & 0.5026 & 5.6\% \\ \hline
Pythia-70M & Pythia-1.4B & \multicolumn{1}{l|}{0.5026} & 0.5153 & 2.5\% & \multicolumn{1}{l|}{0.4719} & 0.5089 & 7.8\% \\ \hline
Pythia-70M & Pythia-2.8B & \multicolumn{1}{l|}{0.5} & 0.5115 & 2.3\% & \multicolumn{1}{l|}{0.4923} & 0.4987 & 1.3\% \\ \hline
Pythia-160M & Pythia-410M & \multicolumn{1}{l|}{0.5} & 0.5281 & 5.6\% & \multicolumn{1}{l|}{0.4758} & 0.4834 & 1.6\% \\ \hline
Pythia-160M & Pythia-1B & \multicolumn{1}{l|}{0.5026} & 0.523 & 4.1\% & \multicolumn{1}{l|}{0.4681} & 0.4898 & 4.6\% \\ \hline
Pythia-160M & Pythia-1.4B & \multicolumn{1}{l|}{0.5077} & 0.5217 & 2.8\% & \multicolumn{1}{l|}{0.4936} & 0.5038 & 2.1\% \\ \hline
Pythia-160M & Pythia-2.8B & \multicolumn{1}{l|}{0.5077} & 0.5153 & 1.5\% & \multicolumn{1}{l|}{0.4949} & 0.5128 & 3.6\% \\ \hline
Pythia-410M & Pythia-1B & \multicolumn{1}{l|}{0.551} & 0.551 & 0\% & \multicolumn{1}{l|}{0.4921} & 0.5179 & 5.2\% \\ \hline
Pythia-410M & Pythia-1.4B & \multicolumn{1}{l|}{0.5816} & 0.5918 & 1.8\% & \multicolumn{1}{l|}{0.5268} & 0.537 & 1.9\% \\ \hline
Pythia-410M & Pythia-2.8B & \multicolumn{1}{l|}{0.5599} & 0.611 & 9.1\% & \multicolumn{1}{l|}{0.5434} & 0.5485 & 0.9\% \\ \hline
Pythia-1B & Pythia-1.4B & \multicolumn{1}{l|}{0.5982} & 0.6288 & 5.1\% & \multicolumn{1}{l|}{0.5536} & 0.574 & 3.7\% \\ \hline
Pythia-1B & Pythia-2.8B & \multicolumn{1}{l|}{0.6288} & 0.6543 & 4.1\% & \multicolumn{1}{l|}{0.5855} & 0.5957 & 1.7\% \\ \hline
Pythia-1.4B & Pythia-2.8B & \multicolumn{1}{l|}{0.6926} & 0.713 & 2.9\% & \multicolumn{1}{l|}{0.6161} & 0.6288 & 2.1\% \\ \hline

\end{tabular}%
}
\caption{This table shows weak to strong generalization using random as well as easy-hard data-splits for quartz dataset. As compared to previous tables \ref{tab:quartz_random} and \ref{tab:quartz_easy}, here we run experiment once and note the improvement of our method with respect to the baseline.}
\label{tab:quartz_improv}
\end{table}

\subsubsection{ Supervised-Fine Tuning task for ARC Question-Answer Dataset}

\begin{table}[H]
\resizebox{\textwidth}{!}{%
\begin{tabular}{llllllll}
                     & \multicolumn{3}{l}{\cellcolor[HTML]{9698ED}Weak Model}  & \cellcolor[HTML]{9698ED} & \multicolumn{3}{l}{\cellcolor[HTML]{9698ED}Strong Model} \\
                     & Token-Avg Acc    & Option Acc     & Option Acc(on w2s)  & $\alpha$                 & oracle         & Token-Avg Acc  & Option Acc             \\
                     & \multicolumn{3}{l}{\cellcolor[HTML]{C0C0C0}Pythia-70m}  & \cellcolor[HTML]{EFEFEF} & \multicolumn{3}{l}{\cellcolor[HTML]{EFEFEF}Pythia-160m}  \\
Baseline             & 13.28 ± 0.05     & 25.31 ± 0.1    & 25.76 ± 0.94        & 10.73 ± 0.03             & 24.12 ± 0.48   & 26.91 ± 0.1    & 24.46 ± 0.06           \\
With Adaboost (T:03) & 17.93 ± 0.78     & 24.75 ± 0.76   & 25.82 ± 0.69        & 10.68 ± 0.02             & 24.12 ± 0.48   & 27.15 ± 0.36   & \textbf{24.23 ± 0.08}  \\
                     & \multicolumn{3}{l}{\cellcolor[HTML]{C0C0C0}Pythia-70m}  & \cellcolor[HTML]{EFEFEF} & \multicolumn{3}{l}{\cellcolor[HTML]{EFEFEF}Pythia-410m}  \\
Baseline             & 13.28 ± 0.05     & 25.31 ± 0.1    & 25.76 ± 0.94        & 10.73 ± 0.03             & 28.61 ± 0.08   & 41.29 ± 0.1    & 27.25 ± 0.24           \\
With Adaboost (T:04) & 17.94 ± 0.88     & 24.97 ± 0.69   & 25.82 ± 0.69        & 10.67 ± 0.04             & 28.61 ± 0.08   & 41.61 ± 0.02   & \textbf{27.27 ± 0.3}   \\
                     & \multicolumn{3}{l}{\cellcolor[HTML]{C0C0C0}Pythia-70m}  & \cellcolor[HTML]{EFEFEF} & \multicolumn{3}{l}{\cellcolor[HTML]{EFEFEF}Pythia-1b}    \\
Baseline             & 13.28 ± 0.05     & 25.31 ± 0.1    & 25.76 ± 0.94        & 10.73 ± 0.03             & 31.11 ± 0.02   & 45.13 ± 0.11   & 28.33 ± 0.18           \\
With Adaboost (T:05) & 19.7 ± 1.18      & 24.92 ± 0.28   & 26.23 ± 0.49        & 10.65 ± 0.04             & 31.11 ± 0.02   & 45.17 ± 0.11   & \textbf{28.52 ± 0.09}  \\
                     & \multicolumn{3}{l}{\cellcolor[HTML]{C0C0C0}Pythia-70m}  & \cellcolor[HTML]{EFEFEF} & \multicolumn{3}{l}{\cellcolor[HTML]{EFEFEF}Pythia-1.4b}  \\
Baseline             & 13.35 ± 0.06     & 25.06 ± 0.14   & 24.39 ± 0.42        & 10.77 ± 0.06             & 32.34 ± 0.3    & 45.21 ± 0.24   & 29.86 ± 0.28           \\
With Adaboost (T:04) & 19.75 ± 1.16     & 24.26 ± 0.56   & 25.7 ± 0.65         & 10.68 ± 0.05             & 32.34 ± 0.3    & 45.33 ± 0.14   & \textbf{30.35 ± 0.13}  \\
                     & \multicolumn{3}{l}{\cellcolor[HTML]{C0C0C0}Pythia-70m}  & \cellcolor[HTML]{EFEFEF} & \multicolumn{3}{l}{\cellcolor[HTML]{EFEFEF}Pythia-2.8b}  \\
Baseline             & 13.42 ± 0.11     & 24.63 ± 0.13   & 23.97 ± 0.55        & 10.77 ± 0.05             & 35.18 ± 0.02   & 48.07 ± 0.12   & 30.94 ± 0.13           \\
With Adaboost (T:02) & 19.88 ± 0.56     & 24.52 ± 0.49   & 24.87 ± 0.81        & 10.68 ± 0.04             & 35.18 ± 0.02   & 47.75 ± 0.08   & \textbf{31.43 ± 0.43}  \\
                     & \multicolumn{3}{l}{\cellcolor[HTML]{C0C0C0}Pythia-160m} & \cellcolor[HTML]{EFEFEF} & \multicolumn{3}{l}{\cellcolor[HTML]{EFEFEF}Pythia-410m}  \\
Baseline             & 25.5 ± 0.66      & 24.12 ± 0.45   & 26.06 ± 0.68        & 9.89 ± 0.03              & 29.18 ± 0.04   & 41.39 ± 0.14   & 27.5 ± 0.27            \\
With Adaboost (T:04) & 31.95 ± 0.47     & 24.94 ± 0.29   & 25.88 ± 0.64        & 9.74 ± 0.03              & 29.18 ± 0.04   & 41.28 ± 0.03   & \textbf{27.7 ± 0.34}   \\
                     & \multicolumn{3}{l}{\cellcolor[HTML]{C0C0C0}Pythia-160m} & \cellcolor[HTML]{EFEFEF} & \multicolumn{3}{l}{\cellcolor[HTML]{EFEFEF}Pythia-1b}    \\
Baseline             & 25.5 ± 0.66      & 24.12 ± 0.45   & 26.06 ± 0.68        & 9.89 ± 0.03              & 31.26 ± 0.44   & 45.12 ± 0.05   & 28.24 ± 0.18           \\
With Adaboost (T:02) & 32.25 ± 0.21     & 24.52 ± 0.34   & 26.06 ± 0.57        & 9.66 ± 0.01              & 31.26 ± 0.44   & 45.18 ± 0.14   & \textbf{28.47 ± 0.24}  \\
                     & \multicolumn{3}{l}{\cellcolor[HTML]{C0C0C0}Pythia-160m} & \cellcolor[HTML]{EFEFEF} & \multicolumn{3}{l}{\cellcolor[HTML]{EFEFEF}Pythia-1.4b}  \\
Baseline             & 24.74 ± 0.14     & 23.97 ± 0.36   & 25.76 ± 0.51        & 9.86 ± 0.02              & 32.25 ± 0.35   & 45.01 ± 0.1    & 30.55 ± 0.07           \\
With Adaboost (T:03) & 32.55 ± 0.21     & 24.46 ± 0.22   & 26.12 ± 0.8         & 9.66 ± 0.01              & 32.25 ± 0.35   & 45.23 ± 0.05   & \textbf{30.86 ± 0.33}  \\
                     & \multicolumn{3}{l}{\cellcolor[HTML]{C0C0C0}Pythia-160m} & \cellcolor[HTML]{EFEFEF} & \multicolumn{3}{l}{\cellcolor[HTML]{EFEFEF}Pythia-2.8b}  \\
Baseline             & 25.43 ± 0.66     & 24.34 ± 0.09   & 26.0 ± 0.32         & 9.86 ± 0.02              & 35.44 ± 0.06   & 47.88 ± 0.02   & 31.03 ± 0.15           \\
With Adaboost (T:04) & 32.6 ± 0.03      & 24.23 ± 0.18   & 26.47 ± 0.53        & 9.66 ± 0.02              & 35.44 ± 0.06   & 47.77 ± 0.08   & \textbf{31.68 ± 0.41}  \\
                     & \multicolumn{3}{l}{\cellcolor[HTML]{C0C0C0}Pythia-410m} & \cellcolor[HTML]{EFEFEF} & \multicolumn{3}{l}{\cellcolor[HTML]{EFEFEF}Pythia-1b}    \\
Baseline             & 39.76 ± 0.3      & 27.85 ± 0.52   & 24.33 ± 0.97        & 9.39 ± 0.02              & 30.97 ± 0.08   & 44.94 ± 0.08   & 28.9 ± 0.12            \\
With Adaboost (T:02) & 40.69 ± 0.14     & 28.27 ± 0.11   & 24.33 ± 0.59        & 9.01 ± 0.04              & 30.97 ± 0.08   & 44.76 ± 0.14   & \textbf{29.41 ± 0.08}  \\
                     & \multicolumn{3}{l}{\cellcolor[HTML]{C0C0C0}Pythia-410m} & \cellcolor[HTML]{EFEFEF} & \multicolumn{3}{l}{\cellcolor[HTML]{EFEFEF}Pythia-1.4b}  \\
Baseline             & 39.66 ± 0.22     & 27.82 ± 0.53   & 24.09 ± 0.8         & 9.39 ± 0.02              & 32.82 ± 0.27   & 45.54 ± 0.03   & 30.26 ± 0.56           \\
With Adaboost (T:02) & 40.82 ± 0.13     & 28.9 ± 0.21    & 24.51 ± 0.59        & 9.01 ± 0.04              & 32.82 ± 0.27   & 45.66 ± 0.09   & \textbf{30.94 ± 0.53}  \\
                     & \multicolumn{3}{l}{\cellcolor[HTML]{C0C0C0}Pythia-410m} & \cellcolor[HTML]{EFEFEF} & \multicolumn{3}{l}{\cellcolor[HTML]{EFEFEF}Pythia-2.8b}  \\
Baseline             & 39.57 ± 0.24     & 28.01 ± 0.69   & 24.69 ± 0.44        & 9.39 ± 0.01              & 35.86 ± 0.26   & 48.06 ± 0.15   & 31.15 ± 0.3            \\
With Adaboost (T:04) & 40.56 ± 0.11     & 28.7 ± 0.34    & 25.34 ± 1.12        & 9.03 ± 0.07              & 35.86 ± 0.26   & 48.22 ± 0.12   & \textbf{31.88 ± 0.27}  \\
                     & \multicolumn{3}{l}{\cellcolor[HTML]{C0C0C0}Pythia-1b}   & \cellcolor[HTML]{EFEFEF} & \multicolumn{3}{l}{\cellcolor[HTML]{EFEFEF}Pythia-1.4b}  \\
Baseline             & 42.31 ± 0.2      & 30.35 ± 0.24   & 28.02 ± 0.76        & 9.53 ± 0.02              & 32.65 ± 0.43   & 45.41 ± 0.06   & 30.26 ± 0.22           \\
With Adaboost (T:03) & 43.22 ± 0.13     & 31.68 ± 0.55   & 27.79 ± 0.71        & 9.37 ± 0.01              & 32.65 ± 0.43   & 45.44 ± 0.06   & \textbf{31.28 ± 0.22}  \\
                     & \multicolumn{3}{l}{\cellcolor[HTML]{C0C0C0}Pythia-1b}   & \cellcolor[HTML]{EFEFEF} & \multicolumn{3}{l}{\cellcolor[HTML]{EFEFEF}Pythia-2.8b}  \\
Baseline             & 42.2 ± 0.29      & 30.46 ± 0.16   & 27.73 ± 0.89        & 9.53 ± 0.02              & 35.12 ± 0.26   & 48.12 ± 0.06   & 32.14 ± 0.02           \\
With Adaboost (T:02) & 43.61 ± 0.2      & 31.17 ± 0.93   & 27.79 ± 0.76        & 9.26 ± 0.02              & 35.12 ± 0.26   & 48.2 ± 0.08    & \textbf{32.54 ± 0.08}  \\
                     & \multicolumn{3}{l}{\cellcolor[HTML]{C0C0C0}Pythia-1.4b} & \cellcolor[HTML]{EFEFEF} & \multicolumn{3}{l}{\cellcolor[HTML]{EFEFEF}Pythia-2.8b}  \\
Baseline             & 42.39 ± 0.37     & 33.42 ± 0.37   & 30.65 ± 1.82        & 9.48 ± 0.03              & 35.12 ± 0.26   & 48.35 ± 0.11   & 32.42 ± 0.44           \\
With Adaboost (T:02) & 43.58 ± 0.27     & 33.5 ± 0.22    & 30.71 ± 1.48        & 11.07 ± 0.84             & 35.12 ± 0.26   & 48.29 ± 0.13   & \textbf{33.19 ± 0.24} 
\end{tabular}%
}
\caption{This table shows weak to strong generalization using random data-splits for arc dataset. We also study the impact of using ensemble learning methods, which combines weak learners, for weak to strong training. Each model is trained for 5 epochs and uses a learning rate of $5x10^{-5}$. The values in this table are generated by aggregating 3 experiments. We show here mean and Standard Error of the Mean values.}
\label{tab:arc_random}
\end{table}

\begin{figure}[htp]
 \centering
  \includegraphics[trim={0 0 0 0},clip,width=1\linewidth]{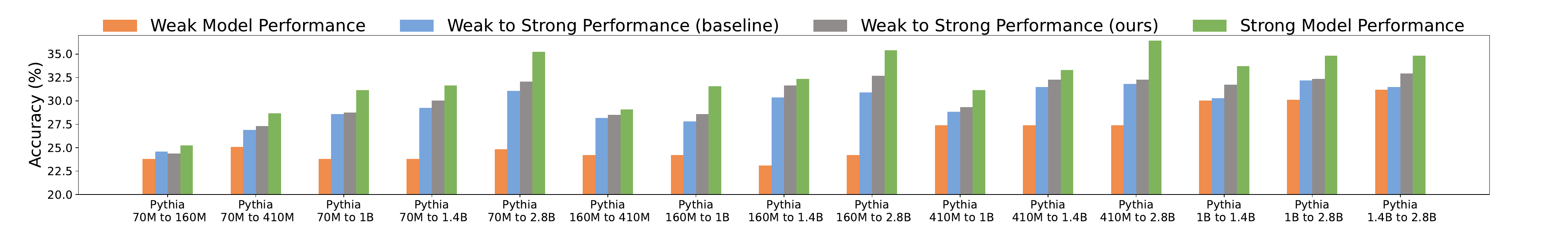}
\caption{\textbf{ARC Dataset (Random):} This figure shows bar plots comparing accuracy values of weak model performance, w2s model performance (baseline and ours) and strong model performance (oracle) for one specific run of experiments. Values are also mentioned in table \ref{tab:arc_easy} and \ref{tab:arc_random}.}
\label{fig:bar_random_arc}
\end{figure}

\begin{table}[H]
\resizebox{\textwidth}{!}{%
\begin{tabular}{llllllll}
                     & \multicolumn{3}{l}{\cellcolor[HTML]{9698ED}Weak Model}  & \cellcolor[HTML]{9698ED} & \multicolumn{3}{l}{\cellcolor[HTML]{9698ED}Strong Model} \\
                     & Token-Avg Acc    & Option Acc     & Option Acc(on w2s)  & $\alpha$                 & oracle         & Token-Avg Acc  & Option Acc             \\
                     & \multicolumn{3}{l}{\cellcolor[HTML]{C0C0C0}Pythia-70m}  & \cellcolor[HTML]{EFEFEF} & \multicolumn{3}{l}{\cellcolor[HTML]{EFEFEF}Pythia-160m}  \\
Baseline             & 8.17 ± 0.06      & 22.5 ± 0.33    & 27.85 ± 0.57        & 10.45 ± 0.0              & 22.3 ± 0.16    & 17.88 ± 0.11   & 22.27 ± 0.32           \\
With Adaboost (T:03) & 13.35 ± 0.54     & 22.81 ± 0.29   & 27.78 ± 0.46        & 10.35 ± 0.02             & 22.3 ± 0.16    & 17.87 ± 0.17   & \textbf{22.56 ± 0.06}  \\
                     & \multicolumn{3}{l}{\cellcolor[HTML]{C0C0C0}Pythia-70m}  & \cellcolor[HTML]{EFEFEF} & \multicolumn{3}{l}{\cellcolor[HTML]{EFEFEF}Pythia-410m}  \\
Baseline             & 8.17 ± 0.06      & 22.5 ± 0.33    & 27.85 ± 0.57        & 10.45 ± 0.0              & 19.28 ± 0.15   & 28.92 ± 0.14   & 17.06 ± 0.31           \\
With Adaboost (T:04) & 14.53 ± 0.72     & 22.93 ± 0.17   & 27.96 ± 0.46        & 10.32 ± 0.0              & 19.28 ± 0.15   & 28.84 ± 0.05   & \textbf{18.0 ± 0.07}   \\
                     & \multicolumn{3}{l}{\cellcolor[HTML]{C0C0C0}Pythia-70m}  & \cellcolor[HTML]{EFEFEF} & \multicolumn{3}{l}{\cellcolor[HTML]{EFEFEF}Pythia-1b}    \\
Baseline             & 8.17 ± 0.06      & 22.5 ± 0.33    & 27.85 ± 0.57        & 10.45 ± 0.0              & 21.5 ± 0.24    & 32.05 ± 0.13   & 19.96 ± 0.15           \\
With Adaboost (T:05) & 12.95 ± 0.88     & 22.58 ± 0.38   & 28.03 ± 0.21        & 10.35 ± 0.02             & 21.5 ± 0.24    & 31.84 ± 0.08   & \textbf{20.45 ± 0.06}  \\
                     & \multicolumn{3}{l}{\cellcolor[HTML]{C0C0C0}Pythia-70m}  & \cellcolor[HTML]{EFEFEF} & \multicolumn{3}{l}{\cellcolor[HTML]{EFEFEF}Pythia-1.4b}  \\
Baseline             & 8.23 ± 0.1       & 22.61 ± 0.42   & 27.37 ± 0.42        & 10.45 ± 0.0              & 21.76 ± 0.14   & 32.98 ± 0.04   & 20.45 ± 0.42           \\
With Adaboost (T:04) & 12.65 ± 0.05     & 23.24 ± 0.06   & 28.32 ± 0.76        & 10.33 ± 0.01             & 21.76 ± 0.14   & 32.95 ± 0.17   & \textbf{21.28 ± 0.02}  \\
                     & \multicolumn{3}{l}{\cellcolor[HTML]{C0C0C0}Pythia-70m}  & \cellcolor[HTML]{EFEFEF} & \multicolumn{3}{l}{\cellcolor[HTML]{EFEFEF}Pythia-2.8b}  \\
Baseline             & 8.33 ± 0.1       & 23.24 ± 0.23   & 27.19 ± 0.47        & 10.45 ± 0.0              & 26.59 ± 0.13   & 35.98 ± 0.09   & 22.78 ± 0.51           \\
With Adaboost (T:02) & 14.28 ± 0.15     & 23.26 ± 0.22   & 28.27 ± 0.14        & 10.37 ± 0.01             & 26.59 ± 0.13   & 35.86 ± 0.28   & \textbf{23.15 ± 0.2}   \\
                     & \multicolumn{3}{l}{\cellcolor[HTML]{C0C0C0}Pythia-160m} & \cellcolor[HTML]{EFEFEF} & \multicolumn{3}{l}{\cellcolor[HTML]{EFEFEF}Pythia-410m}  \\
Baseline             & 17.46 ± 0.16     & 21.73 ± 0.35   & 26.95 ± 0.1         & 9.61 ± 0.0               & 19.11 ± 0.37   & 28.8 ± 0.23    & 18.15 ± 0.15           \\
With Adaboost (T:04) & 20.57 ± 0.1      & 22.16 ± 0.2    & 27.19 ± 0.5         & 9.22 ± 0.02              & 19.11 ± 0.37   & 28.9 ± 0.11    & \textbf{18.43 ± 0.04}  \\
                     & \multicolumn{3}{l}{\cellcolor[HTML]{C0C0C0}Pythia-160m} & \cellcolor[HTML]{EFEFEF} & \multicolumn{3}{l}{\cellcolor[HTML]{EFEFEF}Pythia-1b}    \\
Baseline             & 17.46 ± 0.16     & 21.73 ± 0.35   & 26.95 ± 0.1         & 9.61 ± 0.0               & 21.59 ± 0.07   & 32.06 ± 0.06   & 19.65 ± 0.1            \\
With Adaboost (T:02) & 20.47 ± 0.09     & 22.27 ± 0.29   & 27.31 ± 0.51        & 9.24 ± 0.01              & 21.59 ± 0.07   & 32.07 ± 0.12   & \textbf{20.17 ± 0.14}  \\
                     & \multicolumn{3}{l}{\cellcolor[HTML]{C0C0C0}Pythia-160m} & \cellcolor[HTML]{EFEFEF} & \multicolumn{3}{l}{\cellcolor[HTML]{EFEFEF}Pythia-1.4b}  \\
Baseline             & 17.61 ± 0.07     & 22.84 ± 0.58   & 27.79 ± 0.64        & 9.61 ± 0.0               & 22.33 ± 0.34   & 33.11 ± 0.1    & 21.19 ± 0.15           \\
With Adaboost (T:03) & 20.31 ± 0.24     & 22.5 ± 0.36    & 27.79 ± 0.42        & 9.27 ± 0.06              & 22.33 ± 0.34   & 33.01 ± 0.05   & \textbf{21.25 ± 0.28}  \\
                     & \multicolumn{3}{l}{\cellcolor[HTML]{C0C0C0}Pythia-160m} & \cellcolor[HTML]{EFEFEF} & \multicolumn{3}{l}{\cellcolor[HTML]{EFEFEF}Pythia-2.8b}  \\
Baseline             & 17.64 ± 0.06     & 23.09 ± 0.54   & 27.91 ± 0.59        & 9.6 ± 0.01               & 26.82 ± 0.1    & 35.83 ± 0.36   & 22.44 ± 0.11           \\
With Adaboost (T:04) & 20.3 ± 0.19      & 23.01 ± 0.43   & 27.73 ± 0.25        & 9.26 ± 0.06              & 26.82 ± 0.1    & 36.06 ± 0.07   & \textbf{23.35 ± 0.1}   \\
                     & \multicolumn{3}{l}{\cellcolor[HTML]{C0C0C0}Pythia-410m} & \cellcolor[HTML]{EFEFEF} & \multicolumn{3}{l}{\cellcolor[HTML]{EFEFEF}Pythia-1b}    \\
Baseline             & 27.3 ± 0.16      & 18.8 ± 0.21    & 31.01 ± 0.51        & 9.24 ± 0.0               & 21.33 ± 0.04   & 32.06 ± 0.07   & 20.05 ± 0.08           \\
With Adaboost (T:02) & 28.07 ± 0.12     & 18.35 ± 0.21   & 32.2 ± 0.31         & 8.68 ± 0.09              & 21.33 ± 0.04   & 32.36 ± 0.05   & \textbf{20.34 ± 0.06}  \\
                     & \multicolumn{3}{l}{\cellcolor[HTML]{C0C0C0}Pythia-410m} & \cellcolor[HTML]{EFEFEF} & \multicolumn{3}{l}{\cellcolor[HTML]{EFEFEF}Pythia-1.4b}  \\
Baseline             & 27.5 ± 0.14      & 18.54 ± 0.32   & 31.6 ± 0.21         & 9.24 ± 0.0               & 22.36 ± 0.3    & 33.47 ± 0.07   & 21.13 ± 0.1            \\
With Adaboost (T:02) & 28.09 ± 0.08     & 18.17 ± 0.28   & 31.78 ± 0.4         & 8.67 ± 0.09              & 22.36 ± 0.3    & 33.18 ± 0.11   & \textbf{21.47 ± 0.12}  \\
                     & \multicolumn{3}{l}{\cellcolor[HTML]{C0C0C0}Pythia-410m} & \cellcolor[HTML]{EFEFEF} & \multicolumn{3}{l}{\cellcolor[HTML]{EFEFEF}Pythia-2.8b}  \\
Baseline             & 27.48 ± 0.13     & 18.12 ± 0.13   & 31.66 ± 0.17        & 9.25 ± 0.01              & 26.03 ± 0.21   & 36.13 ± 0.09   & 23.07 ± 0.18           \\
With Adaboost (T:04) & 27.96 ± 0.11     & 18.09 ± 0.2    & 31.07 ± 0.27        & 8.69 ± 0.08              & 26.03 ± 0.21   & 35.93 ± 0.09   & \textbf{24.06 ± 0.15}  \\
                     & \multicolumn{3}{l}{\cellcolor[HTML]{C0C0C0}Pythia-1b}   & \cellcolor[HTML]{EFEFEF} & \multicolumn{3}{l}{\cellcolor[HTML]{EFEFEF}Pythia-1.4b}  \\
Baseline             & 30.64 ± 0.17     & 21.22 ± 0.72   & 32.5 ± 0.6          & 9.38 ± 0.01              & 22.01 ± 0.21   & 33.13 ± 0.11   & 21.5 ± 0.07            \\
With Adaboost (T:03) & 30.41 ± 0.42     & 21.11 ± 0.22   & 32.68 ± 0.56        & 10.98 ± 0.78             & 22.01 ± 0.21   & 33.31 ± 0.03   & \textbf{21.53 ± 0.08}  \\
                     & \multicolumn{3}{l}{\cellcolor[HTML]{C0C0C0}Pythia-1b}   & \cellcolor[HTML]{EFEFEF} & \multicolumn{3}{l}{\cellcolor[HTML]{EFEFEF}Pythia-2.8b}  \\
Baseline             & 30.64 ± 0.17     & 21.22 ± 0.72   & 32.5 ± 0.6          & 9.38 ± 0.01              & 25.51 ± 0.2    & 36.14 ± 0.11   & 23.75 ± 0.16           \\
With Adaboost (T:02) & 31.11 ± 0.12     & 21.67 ± 0.18   & 33.21 ± 0.56        & 9.4 ± 0.24               & 25.51 ± 0.2    & 36.13 ± 0.13   & \textbf{23.75 ± 0.06}  \\
                     & \multicolumn{3}{l}{\cellcolor[HTML]{C0C0C0}Pythia-1.4b} & \cellcolor[HTML]{EFEFEF} & \multicolumn{3}{l}{\cellcolor[HTML]{EFEFEF}Pythia-2.8b}  \\
Baseline             & 31.09 ± 0.12     & 22.27 ± 0.55   & 34.05 ± 0.1         & 9.31 ± 0.01              & 25.26 ± 0.11   & 36.13 ± 0.05   & 23.49 ± 0.2            \\
With Adaboost (T:02) & 31.56 ± 0.1      & 21.79 ± 0.44   & 34.35 ± 0.59        & 10.89 ± 0.65             & 25.26 ± 0.11   & 36.36 ± 0.2    & \textbf{24.37 ± 0.16} 
\end{tabular}%
}
\caption{This table shows weak to strong generalization using easy-hard data-splits for ARC dataset. We also study the impact of using ensemble learning methods, which combines weak learners, for weak to strong training. Each model is trained for 5 epochs and uses a learning rate of $5\times10^{-5}$. The values in this table are generated by aggregating 3 experiments. We show here mean and Standard Error of the Mean values.}
\label{tab:arc_easy}
\end{table}

\begin{figure}[H]
 \centering
  \includegraphics[trim={0 0 0 0},clip,width=1\linewidth]{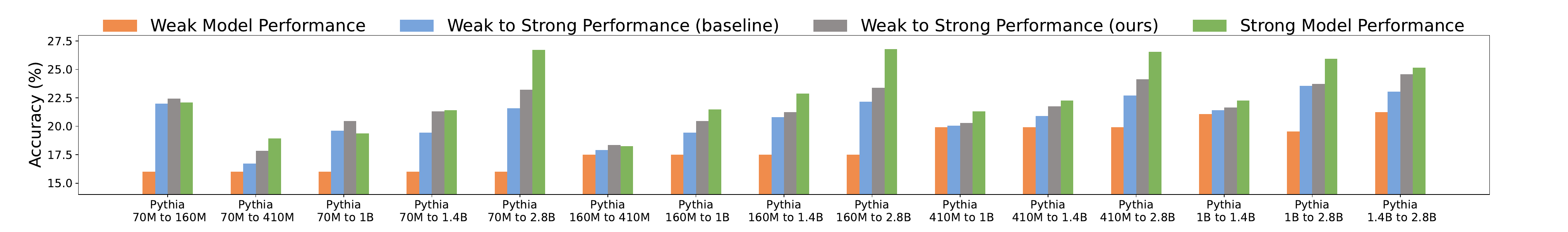}
\caption{\textbf{ARC Dataset (Easy-Hard):} This figure shows bar plots comparing accuracy values of weak model performance, w2s model performance (baseline and ours) and strong model performance (oracle) for one specific run of experiments. Values are also mentioned in table \ref{tab:arc_easy} and \ref{tab:arc_random}.}
\label{fig:bar_easy_arc}
\end{figure}